\documentclass{article}

\usepackage{arxiv}

\usepackage[utf8]{inputenc} % allow utf-8 input
\usepackage[T1]{fontenc}    % use 8-bit T1 fonts
\usepackage{hyperref}       % hyperlinks
\usepackage{url}            % simple URL typesetting
\usepackage{booktabs}       % professional-quality tables
\usepackage{amsfonts}       % blackboard math symbols
\usepackage{nicefrac}       % compact symbols for 1/2, etc.
\usepackage{microtype}      % microtypography
\usepackage{lipsum}		% Can be removed after putting your text content
\usepackage{graphicx}
\usepackage{natbib}
\usepackage{doi}
\usepackage{xcolor}
\usepackage{subcaption}
\usepackage{hyperref}
\usepackage{placeins}
\usepackage{float}

\usepackage{multirow} 
\usepackage{amsmath}
\usepackage{enumitem}
\usepackage{wrapfig}
\usepackage{graphicx}
\usepackage[most]{tcolorbox}
\usepackage{fancyvrb}
\usepackage{fvextra}
\usepackage{tcolorbox}
\tcbuselibrary{breakable} % 启用分页

\title{DeepThink3D: Enhancing Large Language Models with Programmatic Reasoning in Complex 3D Situated Reasoning Tasks}

\author{
 Jiayi Song$^{1,\dagger}$, Rui Wan$^{1,\dagger}$, Lipeng Ma$^1$, Weidong Yang$^{1,}$\thanks{Corresponding authors}, Qingyuan Zhou$^1$, Yixuan Li$^1$, Ben Fei$^{2,}$\protect\footnotemark[1]\\
  $^1$ Fudan University, $^2$ The Chinese University of Hong Kong\\
  \texttt{22307130359@m.fudan.edu.cn, 23210240298@m.fudan.edu.cn, lpma21@m.fudan.edu.cn,} \\
  \texttt{wdyang@fudan.edu.cn, benfei@cuhk.edu.hk}
  }

\date{}

\renewcommand{\shorttitle}{DeepThink3D}

\hypersetup{
pdftitle={A template for the arxiv style},
pdfsubject={q-bio.NC, q-bio.QM},
pdfauthor={David S.~Hippocampus, Elias D.~Striatum},
pdfkeywords={First keyword, Second keyword, More},
}

\begin{document}
\maketitle

\begin{abstract}
This work enhances the ability of large language models (LLMs) to perform complex reasoning in 3D scenes. 
Recent work has addressed the 3D situated reasoning task by invoking tool usage through large language models. 
Large language models call tools via APIs and integrate the generated programs through a chain of thought to solve problems based on the program results.
However, due to the simplicity of the questions in the dataset, the generated program reasoning chains are relatively short. 
To solve this main challenge, in this paper, we introduce DeepThink3D to enhance the tool usage of LLMs in complex 3D situated reasoning tasks.
Our work proposes a combinatorial and iterative evolutionary approach on the SQA3D benchmark to generate more complex questions. 
Building on this foundation, we fine-tune the large language model to make it more proficient in using 3D tools. 
By employing Direct Preference Optimization (DPO), we directly optimize the toolchain strategies generated by models, thereby enhancing their accuracy in complex tasks.
\end{abstract}

% keywords can be removed
% \keywords{First keyword \and Second keyword \and More}

\vspace{-0.3cm}
\section{Introduction}
\label{sec:introduction}

With the rapid advancement of embodied AI~\cite{duan2022survey,liu2024aligning,ma2024survey,feng2025multi}, enabling agents to understand and reason about the 3D physical world has become a critical step toward achieving higher-level cognition and interaction capabilities.
In this context, 3D Situated Reasoning (3D-SR)~\cite{ma2022sqa3d} has emerged as a key research direction, requiring agents to integrate 3D perception, natural language understanding, and spatial reasoning—all from a first-person perspective—to perform complex tasks such as question answering, navigation, and planning.
% For example, in a virtual home environment, an agent needs to interpret the current scene by analyzing visible objects and their spatial relationships to answer questions like ``What is the shortest route to the kitchen from here?''
These capabilities are essential for tasks that require spatial understanding and multi-step reasoning in complex 3D environments, such as embodied navigation and interactive reasoning.
Compared to conventional visual question answering (VQA)~\cite{antol2015vqa,wu2017visual} or purely language-based reasoning~\cite{xiong2023natural}, 3D-SR introduces new demands by emphasizing contextual scene modeling and multimodal reasoning across diverse sources of information.

% In recent years, 3D-SR tasks have drawn increasing attention in applications such as embodied navigation, virtual question answering systems, and augmented reality. 
% These tasks require models to perform spatial understanding and multi-step reasoning in complex 3D environments to complete operations such as navigation, object recognition, and task planning. 

Recent methods~\cite{ma2022sqa3d,radford2021learning,3dllm,huang2023embodied,linghu2024multi,hao2024embosr} have made some progress in 3D Situated Reasoning.
However, mainstream approaches~\cite{ma2022sqa3d,radford2021learning,zhu20233d,man2024situational} typically rely on end-to-end multimodal training, jointly encoding 3D visual scenes and language inputs, and directly predicting answers via supervised learning.
While these models may perform well on in-domain datasets, they exhibit three critical limitations in real-world applications: (1) poor generalization to open environments, (2) opaque decision-making processes with limited interpretability and controllability, and (3) heavy dependence on expensive, high-quality annotated data that constrains scalability.
% face notable challenges in real-world scenarios. 
% Specifically, they struggle to generalize to open environments, operate as black-box systems with limited interpretability and controllability, and rely heavily on high-quality annotated data, which makes training expensive and limits scalability.

% To overcome these limitations, researchers have started to explore integrating the reasoning capabilities of large language models (LLMs).
To address these limitations, recent work~\cite{3dllm,huang2023embodied,qingrong2024llm-tpc,fu2024scene} has investigated incorporating large language models (LLMs) for their reasoning capabilities.
LLMs have demonstrated impressive performance in natural language understanding and commonsense reasoning, presenting new opportunities for 3D-SR.
Current methodologies fall into two main categories: (1) approaches that convert 3D scenes into textual representations for direct LLM-based question answering, and (2) methods that leverage LLMs to generate executable code for interacting with 3D environments through tool APIs, thereby decomposing complex tasks into sequential reasoning steps. 
Notably, these tool-augmented approaches demonstrate significant potential for enhancing both interpretability and complex reasoning capacity.
% These models have demonstrated impressive performance in natural language understanding and commonsense reasoning, offering new avenues for 3D-SR. 
% Some approaches translate 3D scenes into language descriptions and feed them directly to LLMs for question answering; others generate executable code using LLMs to interact with 3D environments via tool APIs, solving reasoning subtasks step by step. 
% These tool-augmented approaches show promise in improving interpretability and enabling more complex reasoning.
\begin{figure*}[t]
    \vspace{-0.3cm}
    \centering
    \includegraphics[width=\columnwidth]{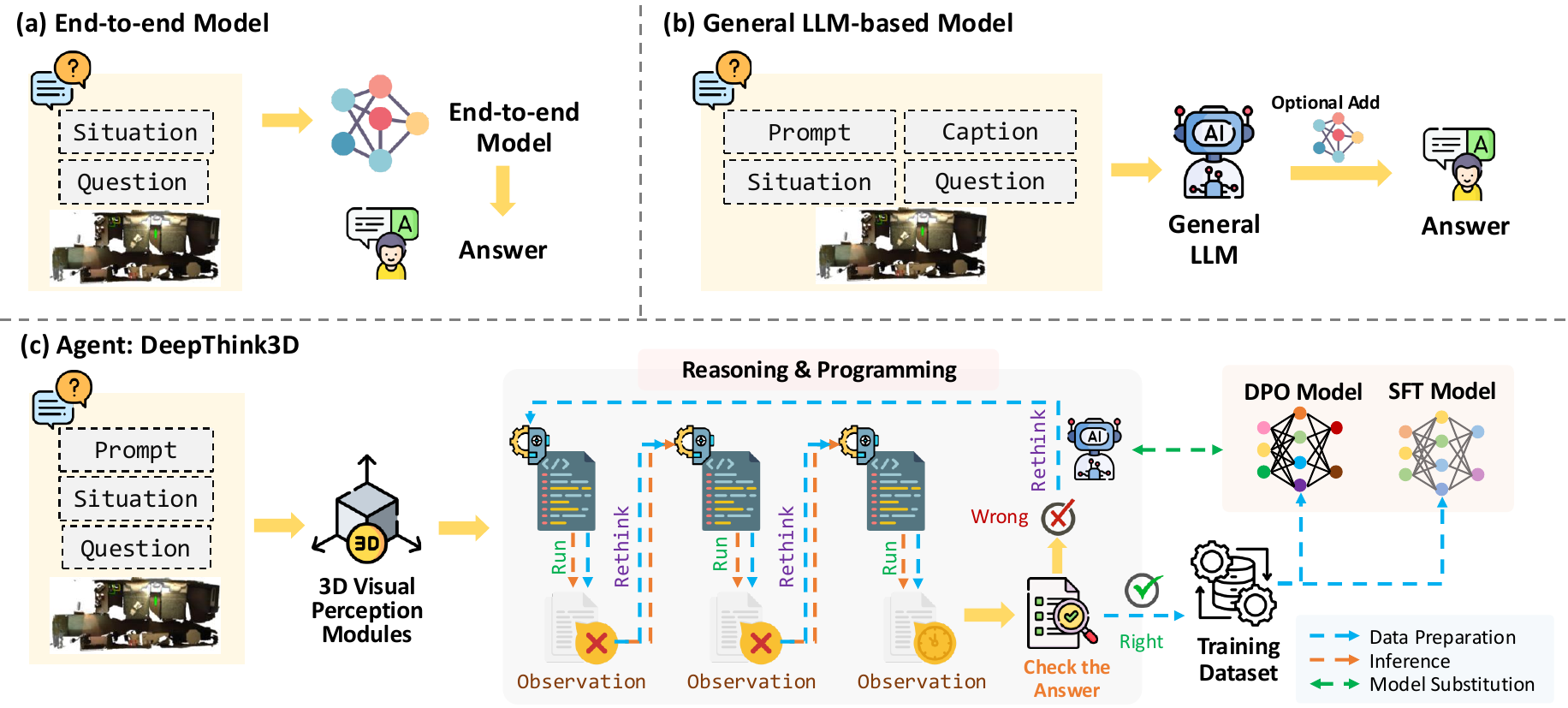}
    % \vspace{-0.7cm}
    \caption{\textbf{Frameworks of Mainstream Methods and DeepThink3D}. End-to-end models lack generalization, interpretability, and depend heavily on annotated data, while genenral LLM-based methods struggle with unclear reasoning and low code executability. Our DeepThink3D addresses these limitations by clarifying the reasoning chain of the LLM and enhancing the executability of the generated code.}
    \label{fig:teaser}
    \vspace{-0.3cm}
\end{figure*}

%基于代码的推理，由于融合了API，并做了问题拆解，优于
% 1. 当前在3d位置推理场景下的推理能力弱 -> 不知道如何调用api解决问题
% 2. 3d场景代码的可执行性，api的引入使得代码的正确性受到了挑战，例如api使用错误；
However, existing models~\cite{3dllm,qingrong2024llm-tpc} based on general-purpose LLMs for 3D-SR tasks still face two key challenges.
One key challenge lies in the weak reasoning ability of these models.
Specifically, they often struggle to distinguish between reasoning and acting behaviors. 
These model often focuses on generating the correct final answer and may neglect or overuse reasoning or acting steps during the process, leading to an unstructured and incorrect reasoning path. 
This results in a lack of coherent decision-making, making it difficult for the model to effectively solve complex tasks.
Another challenge is the limited executable capability of the code generated by LLMs.
Due to the inherent complexity of 3D scene tasks and the issues mentioned in the first challenge, the generated code frequently contains errors that prevent it from running properly.
These errors manifest in two ways: (1) the code may have bugs that cause it to fail completely; (2) even when the code runs, incorrect API calls can still occur leading to misaligned execution results. 
These challenges significantly limit the practical use of LLMs for complex tasks in 3D-SR, underscoring the need for stronger mechanisms to ensure the accuracy and functionality of the generated code.
% However, current methods often lack explicit mechanisms to optimize the structure of the toolchain itself, which is crucial for ensuring the success of complex reasoning tasks. 
% The toolchain refers to the sequence of operations or steps the model takes in response to a given problem, and each step must be logically coherent and error-free for the overall task to succeed. 
% The toolchain, defined as the sequence of operations a model executes to solve a problem, requires strict logical coherence at each step to maintain accuracy. 
% Unfortunately, even a single erroneous or misaligned step can trigger cascading failures, compromising the entire reasoning process. 
% This lack of error tolerance highlights a significant weakness in the model's robustness and controllability. 
% Consequently, minor flaws in the reasoning chain may lead to significant performance degradation, undermining the system’s reliability for real-world applications where precision is essential.
% Unfortunately, if a single step in the generated program is incorrect or misaligned with the required reasoning, it can cause a cascading failure, rendering the entire reasoning process ineffective. 
% As a result, even minor flaws in the reasoning chain can lead to poor performance, making it difficult to trust the system for real-world applications where precision and reliability are essential.

% 强调一下think的过程，think+code的形式做的中间过程。
% 代码本身就存在问题：1. 代码SFT激发思维的内容； 2. 由于API的引入，代码的正确性（正确运行&代码运行正确）受到了挑战。

To address these challenges, we propose DeepThink3D, a framework designed to improve LLMs’ ability to reason about 3D environments through structured tool use. 
Our key motivation is that complex reasoning tasks demand not only the correct selection of tools but also a systematic approach to composing them into logically coherent and executable toolchains.
To achieve this,  we introduce a two-stage optimization approach to systematically enhance code generation:
% First, we apply Supervised Fine-Tuning (SFT) by using programs derived from reasoning chains generated by Large Language Models (LLMs) on the SQA3D dataset as the training data. 
First, we apply Supervised Fine-Tuning (SFT)~\cite{zhang2023instruction} by using programs derived from LLM-generated reasoning chains on the SQA3D dataset as training data.
These programs provide an explicit decomposition of the model’s reasoning process, offering valuable insights into its decision-making steps. 
% These programs offer a clear and explicit breakdown of the underlying reasoning process within the LLMs, effectively providing insight into how the model reaches its conclusions. 
Through this SFT process, the model not only learns to translate natural language questions into interpretable, executable code but also gains the ability to transform what would otherwise be a black-box reasoning process into a more modular, structured, and transparent execution pipeline. 
% This step enhances the model's interpretability and control, as each sub-task in the reasoning chain becomes a distinct, understandable module, allowing for better debugging and fine-tuning of the model’s performance on complex 3D Situated Reasoning (3D-SR) tasks.
This step significantly improves model interpretability and control, as each sub-task in the reasoning chain is rendered as a distinct, debuggable module. 
Consequently, the framework enables more effective performance tuning and error diagnosis in complex 3D-SR tasks.
Building on this, we further introduce Direct Preference Optimization (DPO)~\cite{rafailov2023direct} to refine the model’s toolchain generation capability.
Specifically, we construct DPO training pairs by treating the erroneous program code generated during iterative reasoning as rejected data and the final correct code as chosen data
% Specifically, we treat the problematic program code generated during the iterative reasoning process and the final, correct code as the rejected and chosen training data for DPO, respectively. 
This allows the model to learn from both successful and unsuccessful attempts, enhancing its ability to select more reliable and effective reasoning paths.
Unlike reward modeling or reinforcement learning, DPO directly compares pairs of programs, optimizing the model toward those that demonstrate better structure and higher success rates.  
This process improves the model’s reasoning quality and robustness, further enhancing its ability to generate more reliable and effective toolchains.
Moreover, since the reasoning steps in the SQA3D dataset are relatively simple, training solely on its generated data is insufficient for the model to learn and understand the reasoning process in a deeper level. 
To address this, we used an LLM to combine multiple questions within the same scene from the original SQA3D dataset into more complex, integrated questions, thereby creating more challenging training data to enhance the model's reasoning capabilities.
By explicitly separating perception, reasoning, and execution, DeepThink3D enhances both the controllability and reliability of LLM-based 3D-SR systems, enabling them to tackle complex reasoning tasks in dynamic 3D environments more effectively.

\vspace{-0.3cm}
\section{Related Work}
\label{sec:related}

\textbf{3D Scene Understanding. }
3D scenes serve as representations of the real world; therefore, understanding 3D scenes is a crucial ability for embodied intelligence to interact within the physical environment. 
In recent years, various types of 3D scene tasks have been proposed, such as 3D object caption~\cite{xue2024ulip}, robotic manipulation~\cite{10.5555/3692070.3693552}, task planning~\cite{agia2022taskography,rana2023sayplan}, embodied navigation~\cite{huang2024embodied,3dllm}, and 3D scene question answering~\cite{dai2017scannet}, etc. 
Among them, 3D scene question answering requires the model to generate correct answers to questions in a specific 3D scene. 
The models proposed for 3D question answering are evaluated to identify objects in the scene, extract features, and perform more complex tasks of subsequent reasoning. 
As a type of 3D scene question answering, situated question answering focuses on the characteristics of spatial relationships, requiring the model to reason about spatial problems based on known locations. 
Recent researches release new benchmarks to incorporate visual reasoning questions to 3D data~\cite{linghu2024multi,ma2022sqa3d}. 
However, these tasks usually contain a simple reasoning chain. 
For example, ``How many tables are in front?'' only involves one time of orientation judgment and object recognition. To test the ability to understand 3D scenes further, it is necessary to construct more complex scene problems.

\textbf{LLM for Scene Understanding. }
With the growing capabilities of LLMs, their applications now extend to 3D scene understanding~\cite{ma2024llms,li2025embodied}. 
By integrating visual modules and aligning 3D features with text, models like 3D-LLM~\cite{3dllm} and LEO~\cite{huang2024embodied} use CLIP~\cite{Radford2021LearningTV} and point cloud backbones~\cite{qi2017pointnetplusplus, yu2021pointbert} to extract visual features, which are then fine-tuned on LLMs. 
However, these approaches rely on large, high-quality datasets and struggle to generalize to unseen or noisy scenes.

Alternatively, methods such as~\cite{yang2023mm, surismenon2023vipergpt, qingrong2024llm-tpc} interact with 3D scenes via APIs instead of directly encoding scene features. 
This enables reasoning based on API responses, offering better interpretability and adaptability without additional training. 
Inspired by ViperGPT~\cite{surismenon2023vipergpt}, which generates Python code to compose vision-language modules, LLM-TPC~\cite{qingrong2024llm-tpc} extends this idea to 3D scene understanding tasks like segmentation, attribute classification, and spatial relation recognition.
While LLM-TPC uses the Think process to decompose problems, it still struggle with the understanding of step-by-step reasoning. Its Rectify phase mainly fixes code errors rather than improving reasoning, and the generated code often lacks strong executability.

% Our approach addresses these limitations by incorporating SFT and DPO into the code generation reasoning process.
% SFT enables the model to learn more structured and modular reasoning by training on code generated from iterative reasoning chains, improving its ability to decompose complex tasks into clearer steps.
% DPO further refines this process by comparing successful and failed codes during training, encouraging the model to prefer more effective reasoning paths.
Our approach further trains the model with SFT and DPO to address these issues.
SFT helps the model understand the logic of reasoning, enabling more structured and modular thinking.
DPO optimizes the model’s ability to select more effective reasoning paths and improves code executability.
% Our approach can address these limitations by incorporating Supervised Fine-Tuning (SFT) and Direct Preference Optimization (DPO) specifically for the code generation reasoning process.
% We enhance the model's ability to handle complex reasoning in different tasks by fine-tuning on the code generated from the iterative reasoning chain as training data.
% Through SFT, the model learns how to convert complex natural language questions into clear, executable code, helping break down the reasoning chain into smaller, independent modules.
% Each module is a flexible unit that can be adjusted based on the task's requirements, thus improving the flexibility in handling complex or dynamic tasks.
% Unlike the relatively static toolchain structure in LLM-TPC, our SFT optimization process makes the toolchain calls more flexible and adaptive.
% Besides, DPO further improves the model’s ability to adapt in dynamic task environments.
% In DPO, we not only optimize for the correct toolchains but also learn from failed ones by comparing and contrasting them.
% This comparison-driven training allows the model to adjust the reasoning process when errors occur, selecting a more appropriate toolchain or path to prevent cascading failures.
% DPO enhances the model’s robustness and flexibility by enabling it to quickly adjust its reasoning paths in response to feedback and changing conditions.

\textbf{LLM Programming with APIs. }
LLMs enhance their applications by combining their intrinsic logical reasoning capabilities with additional tools such as APIs~\cite{nam2024using,liu2024autofeedback,qu2025tool}. One approach involves prompting the model with function calls that indicate the available tools. The model then selects the tool and its parameters after deliberation and further refines its response based on the return value of the tool~\cite{chen2024functioncalling}. More advanced, the model can integrate different APIs to form a program, allowing the external execution of more complex logic code. This process tests both the model's understanding of external APIs and its own programming abilities.

% However, several challenges arise when incorporating external APIs into LLM-based programming.
% One key issue is the accuracy and reliability of tool calls. The model's reasoning heavily depends on the APIs' performance. If an API call is incorrect or returns unstable responses, it can disrupt the entire process.
% In addition, error handling remains a concern. Errors in tool calls can affect the entire reasoning process. While some approaches attempt to correct mistakes, they often lack a comprehensive error-prevention mechanism, impacting the model’s robustness.
% Moreover, many methods suffer from a lack of dynamic adaptability. With most approaches using a static toolchain, the model struggles to adjust tools or parameters in response to task complexity. This limits its flexibility, particularly in dynamic or uncertain environments.
However, incorporating external APIs into LLM-based programming presents several challenges.
A key issue is the accuracy and reliability of tool calls. The model’s reasoning heavily depends on API performance, and incorrect calls or unstable responses can disrupt the entire process. 
Additionally, error handling remains difficult. Errors in tool calls can affect the reasoning process, and although some methods try to fix mistakes, they often lack comprehensive error prevention mechanisms, limiting the model’s robustness.

% Although the Rectify strategy in the aforementioned LLM-TPC~\cite{qingrong2024llm-tpc} can help reduce program errors to some extent and provide some flexibility in toolchain calls, its impact is limited. The simple and constrained iterative process it uses can only address the issue to a certain degree, and it does not fully resolve the underlying problems.
Although the Rectify strategy in LLM-TPC~\cite{qingrong2024llm-tpc} helps reduce program errors to some extent, its limited and simplistic iterative process cannot fully address the core issues.
In contrast, our approach fundamentally enhances the interpretability and accuracy of LLMs in generating API-call code through further training. Compared to the monotonous iterative process, our method is better equipped to thoroughly overcome these challenges.

% ~\cite{liu2024autofeedback}

\vspace{-0.3cm}
\section{Method}
\label{sec:method}
\vspace{-0.3cm}
% In this section, we first present an overview of our approach in Sec.\ref{subsec:overview}, which enables LLMs to solve 3D situated reasoning problems through programmatic interaction with external APIs.
In this section, we first introduce the 3D perception modules and APIs employed by our model in the preliminary (Sec.~\ref{subsec:pre}).
We then describe the overall pipeline of our model in Sec.~\ref{subsec:pipeline}, which allows LLMs to tackle 3D situated reasoning tasks via programmatic interaction with external APIs.
After establishing this foundation, we detail how we enhance the model's performance by applying Supervised Fine-tuning and Direct Preference Optimization in Sec.~\ref{subsec:sft} and Sec.~\ref{subsec:dpo}, respectively. 
Additionally, we further elaborated on the method of using LLM to augment the SQA3D dataset in Sec.~\ref{subsec: data}.
% We then introduce how the model generates and executes code in a structured reasoning process guided by tool-augmented API calls in Sec.\ref{subsec:api}.  

\subsection{Preliminary}
\label{subsec:pre}
% In our model, the preprocessing of 3D scenes and the APIs invoked by the LLMs follow the implementation and configuration defined in LLM-TPC~\cite{qingrong2024llm-tpc}.
In our model, the preprocessing of 3D scenes and the APIs invoked by the LLMs follow established implementations and configurations from prior work~\cite{qingrong2024llm-tpc}.
Given a 3D scene, a set of 3D visual perception modules is used to extract structured information from it.
\textbf{3D object segmentation} is carried out using Mask3D~\cite{schult2023mask3d}, which processes point cloud scans to generate object-level instances.
\textbf{Object category and attribute classification} is handled by OpenShape~\cite{liu2023openshape}. By comparing the 3D embedding of each object with textual candidates in the CLIP~\cite{radford2021learning} space, the most semantically relevant label is assigned. 
\textbf{Spatial relation recognition} further identifies geometric relationships between object pairs, including horizontal (e.g., closest, farthest), vertical (e.g., above, on), and allocentric (e.g., left, right) types. These relations are determined based on predefined thresholds over distance, Intersection-over-Union (IoU), and occupancy metrics. 

Based on the 3D visual perception modules mentioned above, LLMs are equipped with four main types of APIs to handle and query object information within 3D scenes: Scene Description (SD), Object Filtering (OF), Object Querying by Relation (OQR), and Object Information Querying (OIQ).

\begin{itemize}[leftmargin=*]
\vspace{-0.3cm}
\setlength{\itemsep}{0pt}
\setlength{\parskip}{0pt}
\item[-] \textbf{Scene Description} (SD) API offers comprehensive information about all objects in the 3D scene, including their positions, attributes, and more. By calling the $scene()$ API, LLMs can obtain detailed data about all objects, providing a full overview of the elements within the scene.
\item[-] \textbf{Object Filtering} (OF) API allows filtering of input objects based on their category. The $filter(x, c)$ API takes an object set $x$ and a category $c$, returning a set of objects $y$ that belong to the specified category. This API facilitates efficient retrieval of specific object types in complex scenes, such as isolating all ``vehicles'' or ``buildings.''
\item[-] \textbf{Object Querying by Relation} (QQR) API is used to find objects that satisfy specific relations with a target object or an agent. By calling $relate(x^t, x^r, r)$, it returns a set of objects $y$ related to the reference object $x^r$ by the relation $r$. The $relate\_agent(x^t, r)$ API, on the other hand, returns objects related to the agent by the relation $r$. These APIs help identify objects in spatial or functional relationships with a specific object or agent, such as finding all objects ``next to'' a target or interacting with an agent.
\item[-] \textbf{Object Information Querying} (QIQ) API provides detailed information about an object's attributes or relations. 
The $query\_relation(x^t, x^r)$ API returns a list of relations between object $x^t$ and reference object $x^r$, while $query\_relation\_agent(x^t)$ returns relations between $xt$ and the agent. The $query\_attribute(x, a_{type} [, a_{cands}])$ API allows querying a specific attribute (e.g., color, shape) of object $x$ and its value, or selecting the most appropriate value from a set of candidates that best matches the object's characteristics.
\end{itemize}

\begin{figure*}[t]
    % \vspace{-1.0cm}
    \centering
    \includegraphics[width=\linewidth]{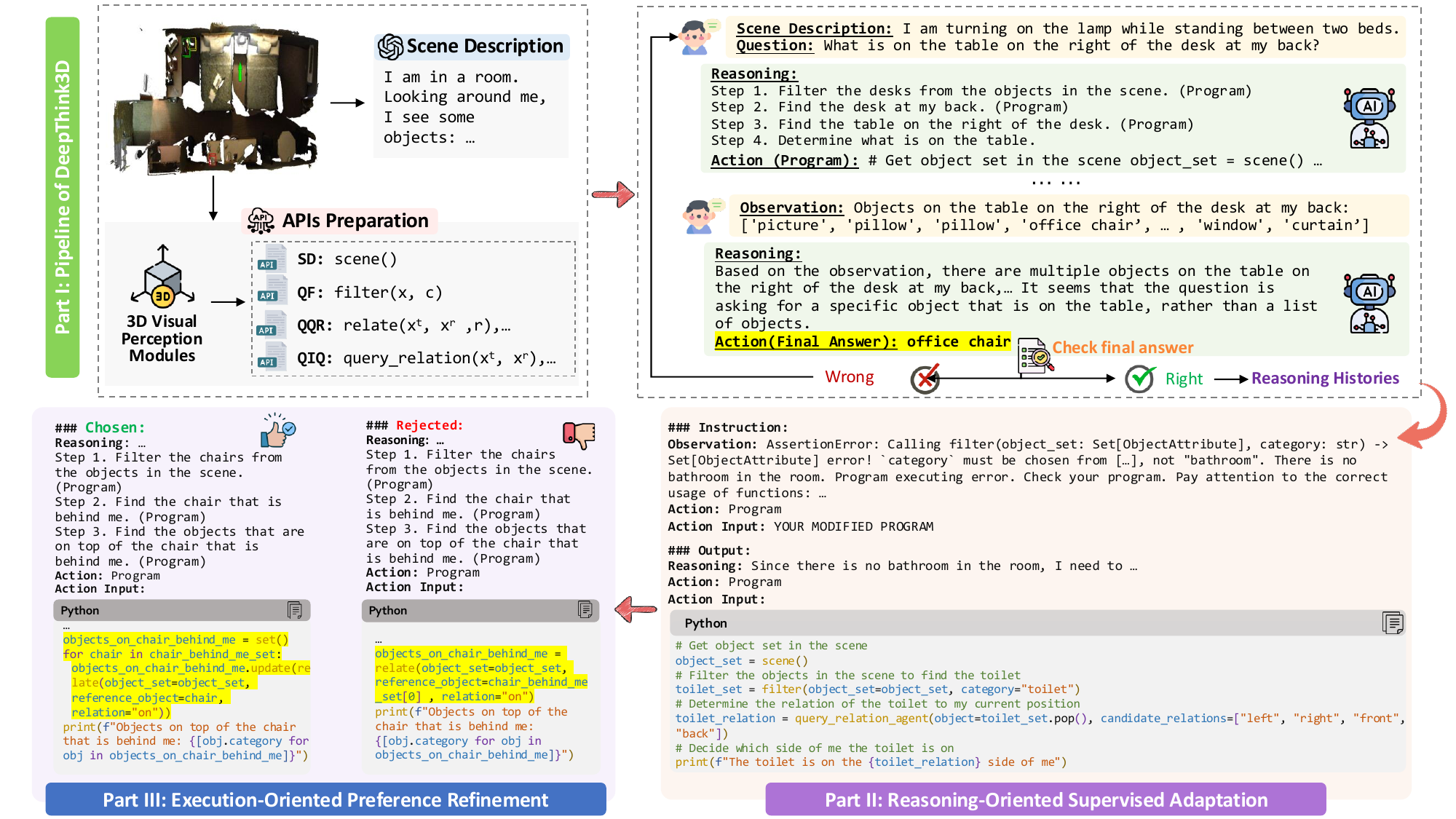}
    % \vspace{-0.7cm}
    \caption{\textbf{Pipeline of DeepThink3D.} Our model generates training data for 3D-SR tasks through a two-layer reasoning and code generation loop. The generated reasoning history is used as fine-tuning data to enhance the model's step-by-step reasoning capability. And both the successful and failed examples of reasoning and code are utilized in preference optimization training to improve the executability of code generated by the LLM.}
    \label{fig:teaser}
    % \vspace{-0.8cm}
\end{figure*}
\subsection{Reasoning-Driven LLM API Coding with Refinement }
\label{subsec:pipeline}
% Our method treats 3D situated reasoning as a learnable code generation task, where a large language model (LLM) generates executable programs that call structured APIs to interact with a 3D environment.
With the APIs mentioned above, our method treats 3D situated reasoning as a learnable code generation task, where an LLM generates executable programs to interact with the 3D environment through structured API calls.
Taking a 3D point cloud of the scene as input, our model first employs 3D visual perception modules to analyze and decompose the environment, producing a representation of the detected object categories. 
This representation, along with the situational context, task question, and API documentation, is then integrated into a prompt for the LLM.
The LLM then performs reasoning over the 3D-SR question, breaking down the complex query into a sequence of intermediate reasoning steps. 
Each step focuses on a specific subgoal, such as identifying relevant objects, inferring spatial relationships, or planning navigational actions.
These steps are then translated into executable code, which interacts with the 3D scene via API calls to extract required information.

Next, the process enters a two-layer loop for code refinement, which is used to generate training data for subsequent SFT and DPO stages.
In this phase, the user executes the code generated by the LLM and feeds the results back to the model.
If the code fails to run correctly due to errors, the output along with the error message is returned to the LLM to reanalyze the failure based on the situation and regenerate the code accordingly. 
This loop continues iteratively until the model produces code that runs successfully or until a predefined maximum number of iteration attempts is reached.
It is worth noting that during the evaluation phase, the model does not have access to the ground-truth answers. Therefore, only a single loop is used at test time, where the model generates an answer to the question based solely on the result of the executed code.

While for the training phase, to collect sufficiently diverse and informative data, we introduce a second loop guided by ground-truth (GT) supervision.
When the model generates code that runs successfully in the first loop, we compare the output of the code with GT answer to the question.
If the answer is correct, the entire code refinement phase terminates.
Otherwise, we inform the LLM that its generated answer is incorrect, prompting it to reconsider the problem, adjust its reasoning, and regenerate the code.
This process repeats until the model generates code that produces the correct answer or until the maximum number of iterations is exceeded. 
This additional loop allows the model to refine its reasoning and code generation based on discrepancies between the generated outputs and the ground-truth answers, thereby enhancing the quality and coverage of the training samples.

The natural language reasoning and code generated during this two-layer loop will be used to create diverse and high-quality training data, which will then serve as the foundation for the subsequent Sec.~\ref{subsec:sft} and Sec.~\ref{subsec:dpo} stages. 
Through this iterative process, the model not only refines its reasoning and code generation but also enhances its ability to handle complex tasks, ensuring that the training data is robust and effective for further optimization.

% Initially, the model generates a plan based on the input, translates it into executable code, and then tests its execution. 
% When the code fails to execuse, the model revises the plan and generates new code, iterating until a correct solution is found.
% Instead of relying solely on trial-and-error iterations, we enhance the model’s reasoning capability through targeted training using Supervised Fine-Tuning (SFT) and Direct Preference Optimization (DPO).
% SFT and DPO are based on the above pipeline. 
% Their training data is constructed by collecting step-by-step codes from the model’s iterative process of code refinement through iterative execution and correction.
% Through SFT and DPO, LLM gains finer control over each step of the iterative reasoning process, effectively reducing errors that may arise during code generation by the LLM.

\subsection{Reasoning-Oriented Supervised Adaptation}
\label{subsec:sft}
To enhance the model's ability to tackle complex tasks through reasoning and iterative correction, we propose a \textbf{Reasoning-Oriented Supervised Fine-tuning} approach. In our approach, the training data for the SFT stage is derived from the iterative reasoning process in multi-round code generation, where the successful paths leading to the correct answer are recorded. In each round, the LLM performs step-by-step reasoning based on the scene description and user query, generates code, executes it, and observes the results. When the output of a particular round of code correctly solves the problem, meaning the results match the ground truth, we record that round and the previous reasoning steps as part of the training sample.

Specifically, each SFT sample can be formalized as a triplet $(q, h, r)$, where:
\begin{itemize}[leftmargin=2em, itemsep=0.1em, topsep=0.1em]
\vspace{-0.1cm}
\setlength{\itemsep}{0pt}
\setlength{\parskip}{0pt}
\item $q$ represents the \textbf{Instruction}, consisting of initial task input and problem (for the first round) or the output and error messages from the code execution in the previous round (for intermediate rounds);
\item $h$ represents the \textbf{History}, including the initial task input and problem as well as the accumulated natural language reasoning and code generated from the first round of iteration up to the previous round;
\item $r$ represents the \textbf{Output}, with the reasoning content and code generated in the current round.
\vspace{-0.1cm}
\end{itemize}
This structure covers problem understanding, historical context, current reasoning intentions, and specific execution strategies during the problem-solving process, providing effective supervision of the model's multidimensional capabilities. During training, we minimize the following conditional log-likelihood loss function:
\begin{equation}
L_{\text{SFT}} = - \mathbb{E}_{(q,h,r) \sim D_{\text{SFT}}} \left[ \log \pi_{\text{base}}(r \mid q, h) \right],
\end{equation}
where $D_{\text{SFT}}$ represents the training sample set extracted from all successful paths, and $\pi_{\text{base}}$ is the base LLM before fine-tuning.

This supervisory approach not only trains the model on "how to do it correctly" but also emphasizes "how to reason step-by-step, analyze feedback, and iteratively correct." This enables the model to develop reasoning decomposition ability for complex tasks. By mimicking the multi-round thinking and debugging paths, the model learns to adjust its problem-solving strategy based on intermediate feedback, thereby improving robustness and generalization in complex scenarios.

\subsection{Execution-Oriented Preference Refinement}
\label{subsec:dpo}
In the process of improving the model's code generation capabilities, we introduce the \textbf{Execution-Oriented Direct Preference Optimization} phase, which focuses on refining the model's output to maximize its ability to generate executable and correct code. 
% Building upon the enhanced reasoning ability acquired during the SFT phase, this stage aims to further improve the LLM's code generation by incorporating feedback from the execution results of the generated programs.
The training data for this phase consists of tuples $(q, r^+, r^-)$, where:
\begin{itemize}[leftmargin=2em, itemsep=0.1em, topsep=0.1em]
\vspace{-0.1cm}
\setlength{\itemsep}{0pt}
\setlength{\parskip}{0pt}
    \item $q$ represents the \textbf{Instruction}, including the task input and problem (similar to the SFT stage);
    \item $r^+$ represents the \textbf{Chosen}, which is the final reasoning and program in the loop that successfully generates the correct answer;
    \item $r^-$ represents the \textbf{Rejected}, which includes two type of programs and their corresponding reasonings: programs that produce errors during execution (i.e., code that cannot run due to errors), and programs that execute correctly but generate incorrect results (i.e., code that runs successfully but outputs the wrong answer).
\vspace{-0.1cm}
\end{itemize}
The goal of this phase is to refine the model's ability to generate executable code by reinforcing the preference for solutions that not only reason correctly but also run without errors and yield correct outputs. 
By utilizing the feedback from both successful and failed executions, the model learns to generate code that is not only logically sound but also practically executable.

The DPO loss function is used to optimize the model during this phase. The loss function seeks to maximize the margin between the likelihood of the correct program and the likelihood of incorrect programs, while maintaining the model's proximity to the reference policy. It is defined as:
\begin{equation}
L_{\text{DPO}} = - \mathbb{E}_{(q,r^+,r^-) \sim D_{DPO}} \left[ \log \sigma \left( \beta \log \frac{\pi_{\text{DPO}}(r^+ \mid q)}{\pi_{\text{SFT}}(r^+ \mid q)} - \beta \log \frac{\pi_{\text{DPO}}(r^- \mid q)}{\pi_{\text{SFT}}(r^- \mid q)} \right) \right],
\end{equation}
where $D_{DPO}$ represents the preference dataset consisting of pairs of correct and incorrect programs, $\sigma(\cdot)$ is the sigmoid function used to model preferences, and $\beta$ is a hyperparameter that controls the strength of the penalty imposed by the KL divergence between the DPO and SFT models. 
$\pi_{\text{DPO}}$ is the policy of the DPO model, while $\pi_{\text{SFT}}$ is the policy of the model after SFT, serving as the reference.

By leveraging this approach, we further enhance the LLM's ability to generate code that not only solves the problem at a logical level but also performs the intended task successfully when executed, effectively enabling it in real-world applications that require accurate and reliable code generation.

\subsection{LLM-Based Question Generation for Data Augmentation}
\label{subsec: data}
Considering that most of the 3D-SR questions in the SQA3D dataset are relatively simple and can typically be answered with less than three straightforward reasoning steps.
To enhance the model’s ability to handle complex reasoning tasks in 3D question answering, we propose an automatic data augmentation strategy based on a large language model (LLM).
The core idea is to generate more logically complex questions on top of the existing dataset, addressing the limited reasoning depth observed in the SQA3D dataset.

Specifically, each sample in SQA3D is associated with a scene identifier (scene\_id), a natural language description of the observation state (situation), and a spatial camera position (position) defined by 3D coordinates and a quaternion rotation. We group questions by their spatial positions within each scene and design a Position class to ensure unique identification and hashing of positions, thereby aggregating all questions posed from the same viewpoint.
After aggregating QA pairs at the position level, we prompt the LLM to generate more challenging questions. Each prompt contains the current observation’s natural language description (situation), the existing question-answer pairs at that position, and an instruction explicitly requesting the model to synthesize a set of new questions with increased semantic complexity and reasoning depth. The generated questions are encouraged to reflect advanced cognitive processes such as multi-step reasoning, entity relation alignment, or spatial position inference, in order to enhance the model’s generalization ability in complex reasoning tasks.

\vspace{-0.3cm}
\section{Experiment}
\label{sec:experiment}

\subsection{Experiment Setup}

We utilize SQA3D~\cite{ma2022sqa3d} as the basis for our experimental benchmark. SQA3D filtered out 650 scenes from the ScanNet dataset~\cite{dai2017scannet} and collected scene recognition and question-answer pairs through manual annotation. The resulting dataset provides the current positional context and requires 3D scene reasoning within this environment. We train and validate our model based on the 26,623 train and 3,519 test samples provided by the SQA3D dataset, and expand the question pairs based on the train set.

In this paper, Llama-3.1-8B-Instruct~\cite{grattafiori2024llama} is selected as the base model for subsequent SFT and DPO stages. In the SFT stage, we train the model for 300 steps on the original data and 1550 steps on the augmented data, setting batch size to 2, gradient accumulation steps to 8, and learning rate to $1.0 \times 10^{-4}$ on 2 NVIDIA A100-SXM4-80GB GPUs. Cosine annealing with linear warm-up is employed, where the warm-up ratio is set to 10\%. 
For the DPO stage, we adjust the learning rate to $5.0 \times 10^{-6}$ and set $\beta$ to 0.1,  training for 100 steps.
Both the SFT and DPO stages employ LoRA fine-tuning~\cite{hu2022lora} with a rank of 8 targeting all layers, and the training process utilizes BF16 precision.

\begin{wraptable}{r}{0.3\textwidth}
% \vspace{-1.5cm}
\caption{Comparison with other methods.}
\label{tab:main}
\centering

\begin{tabular}{lc}
\toprule[1pt]
\multirow{1}{*}{\textbf{Method}}            & \textbf{Acc.}  \\ \midrule 
                                
\multicolumn{1}{l|}{ScanQA~\cite{ma2022sqa3d}}        & 47.74 \\
\multicolumn{1}{l|}{3D-VisTA~\cite{zhu20233d}}      & 50.72 \\
\multicolumn{1}{l|}{3D-LLM~\cite{3dllm}} &  50.21 \\
\multicolumn{1}{l|}{LEO~\cite{huang2023embodied}}           & 53.25 \\
\multicolumn{1}{l|}{LLM-TPC~\cite{qingrong2024llm-tpc}}        &   56.92    \\ \midrule
\multicolumn{1}{l|}{\textbf{DeepThink3D}}  & \textbf{62.11}      \\ \bottomrule[1pt]
\end{tabular}
% \vspace{-0.7cm}

\end{wraptable}

\subsection{Performance Comparison with Existing Approaches}
In this experiment, our method is evaluated against several baseline approaches listed in Table~\ref{tab:main}.
These baselines include both end-to-end multimodal models like ScanQA~\cite{ma2022sqa3d} and 3D-VisTA~\cite{zhu20233d}, as well as general LLM-based methods like 3D-LLM~\cite{3dllm}, LEO~\cite{huang2023embodied} and LLM-TPC~\cite{qingrong2024llm-tpc} that represent the current state-of-the-art in 3D Situated Reasoning. 
% We test their accuracies on the SQA3D dataset~\cite{ma2022sqa3d}.
We evaluate our DeepThink3D on the SQA3D dataset~\cite{ma2022sqa3d}, using accuracy as the primary metric. 
% The metrics of other methods are borrowed from~\cite{qingrong2024llm-tpc}.
As shown in Table~\ref{tab:main}, our DeepThink3D achieves the highest accuracy among all methods, indicating superior reasoning capability and code reliability.

\begin{figure*}[t]
    \vspace{-0.9cm}
    \centering
    \includegraphics[width=\columnwidth]{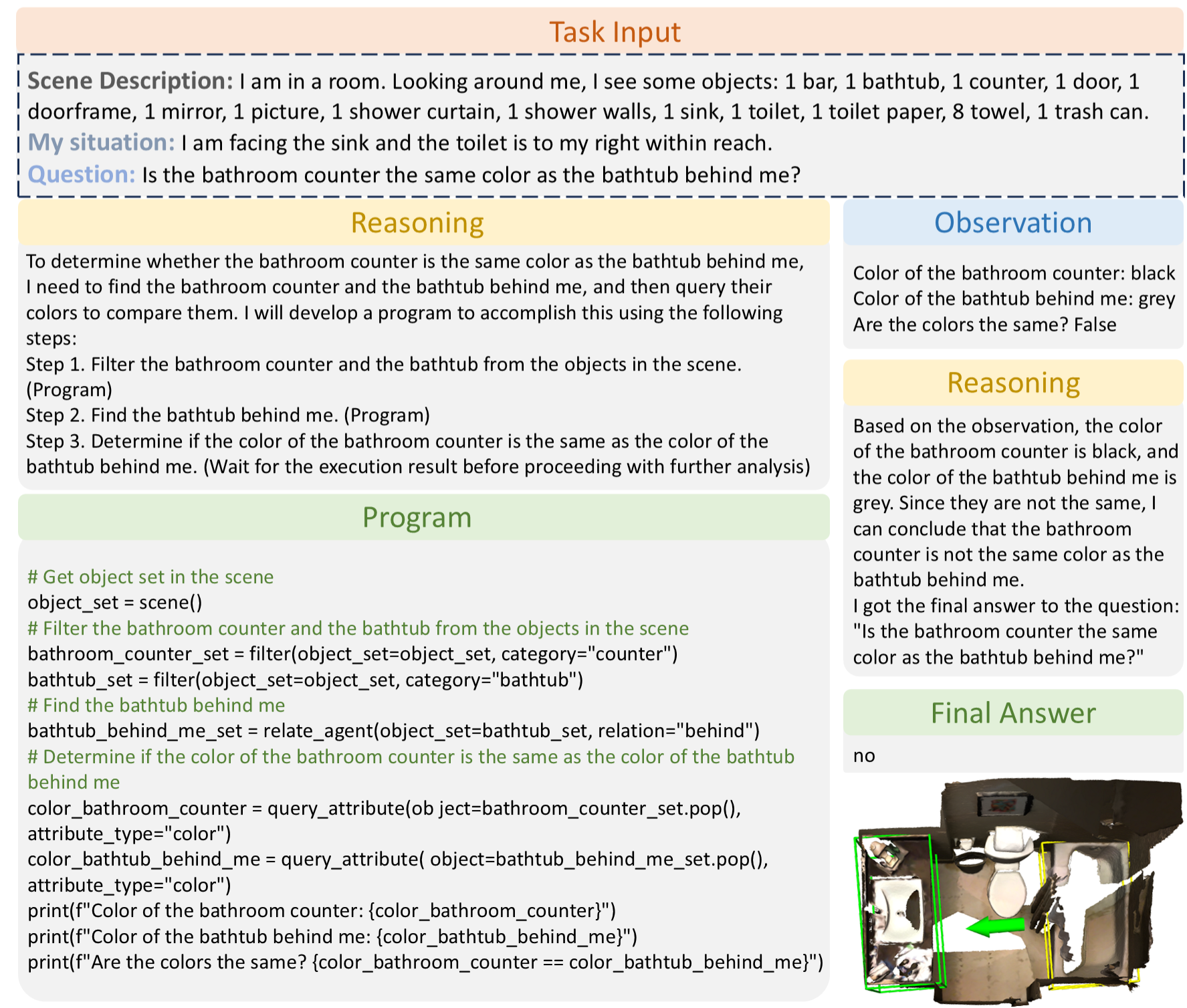}
    % \vspace{-0.7cm}
    \caption{Qualitative result on SQA3D.}
    \label{fig:result}
    % \vspace{-0.7cm}
\end{figure*}

In particular, compared to end-to-end models, our DeepThink3D exhibits stronger interpretability and reasoning transparency, largely due to the explicit reasoning traces and structured execution flow produced during inference, as illustrated in Fig.~\ref{fig:result}. 
This not only helps the model make more coherent decisions but also provides human-readable explanations for its actions.
Compared to existing LLM-based methods, our approach achieves higher execution success due to improved reasoning-code alignment and error-resilient code generation.
Remarkably, \textbf{over 75\%} of the correctly answered questions in the test set are solved with code that returns the correct answer on the \textbf{first} execution, highlighting the improvements brought by our SFT and DPO strategies in enhancing the robustness and efficiency of the reasoning-code generation pipeline.
\vspace{-0.3cm}
\subsection{Ablation Study}
\vspace{-0.1cm}
To better understand the contributions of each component in our framework, we conduct ablation studies on our framework. 
Considering the inherent randomness in the training process, all the following experiments are repeated three times. The reported metrics represent the maximum accuracy and standard deviation across these three runs. 

\begin{wraptable}{r}{0.4\textwidth}
    \vspace{-0.5cm}
    \caption{Ablation Study on SFT and DPO Components.}
    \label{table_ablation_1}
    
    \centering
    \begin{tabular}{lcc}
    \toprule[1pt]
    \multirow{1}{*}{\textbf{Method}} 
     & \textbf{Acc.} & \textbf{Std.} \\
    \midrule
    \multicolumn{1}{l|}{w/o SFT \& DPO}    & 59.83  & 0.027  \\
    \multicolumn{1}{l|}{w/o DPO}           & 58.40 & 0.018 \\
    \multicolumn{1}{l|}{w/o SFT}   & 57.83 & 0.006 \\ \midrule
    \multicolumn{1}{l|}{Ours}     & \textbf{62.11} & 0.035 \\
    \bottomrule[1pt]
    \end{tabular}
    \vspace{-0.5cm}
\end{wraptable}

\textbf{Effect of SFT and DPO Components.}
As shown in Table~\ref{table_ablation_1}, both SFT and DPO contribute significantly to the overall performance of our model. 
Removing either SFT or DPO leads to a noticeable drop in accuracy, indicating that both supervised fine-tuning and preference optimization play complementary roles in enhancing reasoning quality and code executability. 
Compared to the w/o SFT\&DPO setting, applying only SFT or only DPO results in decreased accuracy. This demonstrates that solely learning from supervised fine-tuning without preference-based evaluation, or merely aligning with preferred answers without understanding the underlying reasoning logic, is insufficient. 
Only the combination of learning and preference-based evaluation leads to optimal performance.

\textbf{Impact of DPO Training Data Composition.} 
In our approach, to enhance the executability of the generated code and simultaneously align it more effectively with model’s reasoning ability, we design the rejected samples in the DPO training stage to include two distinct types.
The first type consists of code and corresponding reasoning traces that contain execution errors.
Including these samples helps the model learn to avoid generating code with poor executability.
The second type includes code that executes without errors but produces incorrect outputs due to flawed reasoning. 
These samples are used to refine the model’s reasoning paths and guide it toward invoking the correct APIs based on different task scenarios.

To verify the individual contribution of each type of rejected data, we perform an ablation study on the composition of the DPO training dataset.
Specifically, we compare our full model, trained with both types of rejected samples, against a variant trained using only the first type—samples containing code with execution errors.
In our experiments, the model trained with DPO data containing only a single supervision type achieved an accuracy of \textbf{58.40\%}, with a standard deviation of \textbf{0.017}.
There is a significant drop in performance compared to our model's result of \textbf{62.11\%}, indicating that training solely on data focused on correcting code errors provides limited benefit for the model’s overall understanding of the reasoning process.

\begin{wraptable}{r}{0.55\textwidth}\footnotesize
    \vspace{-0.9cm}
    \caption{Comparison Between Models Trained on Original and Augmented Data.}
    \label{table_ablation_2}
    
    \centering
    \begin{tabular}{lcc|cc}
    \toprule[1pt]
    \multirow{2}{*}{\textbf{Method}}  & \multicolumn{2}{c|}{\textbf{SQA3D}} & \multicolumn{2}{c}{\textbf{SQA3D$_{Ext}$}} \\ \cmidrule{2-5}
     & Acc. & Std. & Acc. & Std. \\
    \midrule
    % LLM-TPC           &  &  &  &  \\
    % \midrule
    % \multicolumn{5}{l}{\textit{Train on SQA3D}} \\
    \multicolumn{1}{l|}{w/o SFT \& DPO}    & 59.83 & 0.027 & 59.83\footnotemark & 0.027\footnotemark[\value{footnote}]  \\
    \multicolumn{1}{l|}{w/o DPO}           & \textbf{60.11} & 0.017 & 58.40 & 0.018 \\ \midrule
    % w/o SFT   &  &  &  &  \\
    \multicolumn{1}{l|}{Ours}     & 58.69 & 0.006 & \textbf{62.11} & 0.035 \\
    % \textit{Ours} on SQA3D$_{Ext}$   &  &  &  &  \\
    % \midrule
    % \multicolumn{5}{l}{\textit{Train on SQA3D$_{Ext}$}} \\
    % DPO without SFT   &  &  &  &  \\
    % DPO After SFT     &  &  &  &  \\
    \bottomrule[1pt]
    \end{tabular}
    \vspace{-0.6cm}
\end{wraptable}

\footnotetext{Since we only perform data augmentation on the training set, the experimental results here are consistent with those on SQA3D.}
\textbf{Effect of Dataset Extension.} 
% The original SQA3D dataset primarily consists of relatively simple questions that can typically be answered through two or three straightforward reasoning steps.
% To better evaluate and enhance our model’s ability to handle more complex reasoning tasks, we extended the dataset by introducing additional questions that require deeper, multi-step inference.
% Specifically, the new QA samples are generated by prompting an LLM to synthesize more complex reasoning questions based on existing ones in the original dataset.
% By leveraging the LLM’s language understanding and generative capabilities, we are able to construct challenging QA instances that preserve the scene context while requiring more elaborate reasoning processes.
% These newly added QA pairs account for 50\% of the original dataset size, thereby significantly increasing the diversity and difficulty of the dataset.
As presented in the Table.~\ref{table_ablation_2}, we compare the performance of models trained with and without data augmentation.
Our model performs significantly better on the augmented dataset compared to before the augmentation.
On the SQA3D dataset, the performance of using only SFT is better than that of SFT+DPO. 
This is because the original dataset contains relatively simple questions, which leads to the model learning limited reasoning logic. 
As a result, DPO introduces biased learning that underperforms compared to the SFT-only version. However, this issue is resolved after incorporating the augmented dataset, which provides more complex and diverse examples for training.

\vspace{-0.3cm}
\section{Limitation}
\label{sec: Limitation}
Despite the promising results, our method still faces some limitations.
(I) First, the effectiveness of our reasoning-code generation framework is influenced by the quality of the visual perception models behind the API calls.
These models handle tasks such as object detection, instance segmentation, and attribute classification in 3D scenes. 
While they provide useful scene understanding, their current capabilities are not perfect and may occasionally produce imprecise outputs.
Such errors, even if small, can affect subsequent reasoning steps and potentially lead to incorrect final results—an issue that is particularly noticeable in 3D Situated Reasoning tasks where intermediate outputs are tightly coupled with downstream reasoning.
(II) Additionally, although our framework leverages powerful LLMs, existing LLMs still exhibit certain limitations.
For instance, the model may sometimes deviate from the predefined instructions or get trapped in local code-level fixes while overlooking the overall reasoning logic.
These factors can introduce unpredictability in the generated solutions, highlighting the necessity of further improvements of LLM to ensure more coherent and reliable reasoning behavior.
% areas where future improvements are needed to ensure more coherent and reliable reasoning behavior.
\vspace{-0.3cm}
\section{Conclusion}
\label{sec:conclusion}
In this work, we present DeepThink3D, a novel framework designed to enhance LLMs' capability in 3D Situated Reasoning tasks through a structured, two-stage reasoning-code generation pipeline.
By decomposing complex tasks into explicit toolchain-based programs, our approach separates the perception, reasoning, and execution stages, improving both interpretability and reliability.
To implement this, we construct a high-quality set of reasoning programs through our two-layer iterative reasoning-coding pipeline and apply SFT to teach the model how to generate interpretable, step-by-step solutions.
Building on this foundation, we further utilize DPO to refine the model’s ability to generate logically coherent and executable toolchains by comparing correct and incorrect code samples during iterative reasoning.
To further enhance model generalization and reasoning capability, we augment the original dataset by synthesizing more complex multi-step questions from simpler ones using LLMs.
This paradigm significantly strengthens the model’s structured reasoning ability and improves the executability of generated code in complex 3D environments.
Experimental results validate the effectiveness of our DeepThink3D, highlighting its potential to serve as a scalable and interpretable solution for embodied reasoning and interaction in 3D environments.
As a result, DeepThink3D demonstrates greater adaptability and robustness in diverse 3D-SR tasks.

% \nolinenumbers

%This is where your bibliography is generated. Make sure that your .bib file is actually called library.bib
\bibliography{ref}

%This defines the bibliographies style. Search online for a list of available styles.
\bibliographystyle{abbrv}
\appendix
\newpage
\clearpage

\section*{Appendix}
% \section{Model Details}
\section{3D Visual Perception Modules}
\subsection{3D Object Segmentation}
The 3D Object Segmentation module in prior work~\cite{qingrong2024llm-tpc} adopts the Mask3D~\cite{schult2023mask3d} model, trained on the ScanNet200~\cite{dai2017scannet} dataset, to perform instance segmentation on scene point clouds. 
While this approach enables fully automated segmentation, ablation studies~\cite{qingrong2024llm-tpc} have shown that predictions from Mask3D lead to a noticeable performance drop compared to ground-truth segmentations, suggesting that current segmentation models still have limitations.
To better highlight the reasoning capabilities of our agent LLM in 3D situated reasoning (3D-SR) tasks, we use ground-truth segmentations to reduce the influence of segmentation noise. 
We look forward to future advances in 3D object segmentation models that can obtain accurate segmentation matching the quality of ground-truth (GT) data, eventually removing the need for manual point cloud annotation in dataset construction.

\subsection{Object Category and Attribute Classification}
In our 3D reasoning pipeline, we integrate both object category and attribute classification modules to provide detailed semantic understanding of 3D scenes.

For object category classification, ground-truth labels corresponding to a comprehensive set of 607 categories consistent with the annotations of SQA3D~\cite{ma2022sqa3d} are utilized. 
These category labels cover common furniture, structural elements, and objects found in indoor scenes, such as ``office chair,'' ``bookshelf,'' and ``ceiling fan.'' 
Each object is linked to its precise category information through its instance ID and the annotation files in the scene, providing accurate semantic labels that support downstream reasoning and question answering tasks. 
This fine-grained category information ensures the semantic coverage required by complex question answering.

For attribute classification, the pre-trained OpenShape~\cite{liu2023openshape} model is employed. 
Specifically, the colored point cloud of each object is fed into OpenShape’s 3D encoder to extract geometric and visual features, while attribute candidates are encoded through the model’s text encoder. 
The cosine similarity between the 3D features and each textual attribute feature is then computed, with the attribute yielding the highest similarity selected as the final prediction. 
This method effectively integrates 3D visual information with textual semantics to achieve accurate attribute recognition.

\subsection{Spatial Relation Recognition}
This module adopts a spatial relation classification scheme derived from the taxonomy used in Nr3D within the ReferIt3D~\cite{achlioptas2020referit3d} benchmark. 
The relation types are categorized into horizontal, vertical, and allocentric relations:
\begin{itemize}
% \vspace{-0.3cm}
% \setlength{\itemsep}{0pt}
% \setlength{\parskip}{0pt}
\item \textbf{Horizontal relations}, such as ``closest,'' ``farthest,'' ``within reach.'' and ``around'';
\item \textbf{Vertical relations}, including ``on.'' ``above,'' and ``below'';
\item \textbf{Allocentric relations}, covering ``left,'' ``right,'' ``front,'' and ``back''.
\end{itemize}
For horizontal relations, spatial proximity is assessed using Euclidean distances between objects. 
The nearest and farthest objects relative to a given target object are determined by sorting all scene objects based on their distances. 
The closest object is identified when the distance to the nearest neighbor is sufficiently smaller (by a margin $\epsilon$) than the next closest. The same criterion applies in reverse to identify the farthest object.
Additional horizontal relations, such as ``within reach'' and ``around'' are defined using distance thresholds. Given a target object and a reference (anchor) object, the target is labeled as ``within reach'' if their distance is below a predefined value 
$wr\_dist$, and as ``around'' if the distance is under another threshold 
$ar\_dist$. These thresholds, along with the margin $\epsilon$, are set as task-specific constants.

In the case of vertical relations, spatial configurations are determined based on both 2D bounding box overlap in the XY plane (i.e., Intersection over Union, IoU) and the height difference along the Z-axis. 
For example, to infer the ``on'' relationship, several conditions must be met: sufficient 2D overlap between the object projections ($IoU>min\_iou$), a vertical gap between the lower surface of the target and the upper surface of the anchor that does not exceed a set threshold, and suitable ratios of overlapping and projected areas. 
These ensure that objects are not only close in vertical position but also spatially aligned in the horizontal plane.

Allocentric relations are handled by spatial reasoning relative to the orientation of the anchor object. 
For instance, determining whether an object is to the ``left'' involves constructing a directional region extending from the left side of the anchor. 
If the target object intersects this region with sufficient area overlap, the ``left'' relation is assigned. 
The same procedure is applied analogously to infer ``right,'' ``front,'' and ``back'' relations.

\section{API Documentation}

Based on the 3D Visual Perception Modules described above, a set of general-purpose APIs is provided for the LLM. 
These APIs are presented in the form of API documentation and are supplied to the LLM along with the task definition prompt at the very beginning.
The detailed API documentation is provided as follows:
\begin{tcolorbox}[colback=gray!10,colframe=gray!80,title=Prompt,breakable]
\begin{Verbatim}[breaklines=true, breaksymbol={}]
When generating a program, each object is represented as an instance of ObjectAttribute and you can use the following functions:
class ObjectAttribute:
    category: str # category of the object
    xyz: List[float] # center coordinates of the object
\end{Verbatim}
\end{tcolorbox}
\subsection{Scene Description (SD)}
\begin{tcolorbox}[colback=gray!10,colframe=gray!80,title=API Documentation, breakable]
\begin{Verbatim}[breaklines=true, breaksymbol={}]
scene() -> Set[ObjectAttribute]:
    """
    Returns a set of objects in the scene.
    """
\end{Verbatim}
\end{tcolorbox}
The \texttt{scene()} function provides a comprehensive snapshot of the current 3D environment by returning a set of all objects present. 
Each object is represented as an instance of \texttt{ObjectAttribute}, which includes information such as the object's category and its spatial coordinates within the scene. 
This function serves as the foundational step for any further operations or queries by supplying the entire collection of objects and their attributes in the environment.
\subsection{Object Filtering (OF)}
\begin{tcolorbox}[colback=gray!10,colframe=gray!80,title=API Documentation, breakable]
\begin{Verbatim}[breaklines=true, breaksymbol={}]
filter(object_set: Set[ObjectAttribute], category: str) -> Set[ObjectAttribute]:
    """
    Returns a set of objects whose category is `category`.

    Examples
    --------
    >>> # Get object set in the scene
    >>> object_set = scene()
    >>> # Filter all the tables
    >>> table_set = filter(object_set=object_set, category="table")
    """
\end{Verbatim}
\end{tcolorbox}
The \texttt{filter()} function enables focused interaction by extracting a subset of objects from the full scene based on their category.
Given an input set of objects and a specified category string, this function returns only those objects whose category matches the input. 
For example, filtering the scene for objects categorized as ``table'' will isolate all tables, allowing subsequent operations to be performed on just this subset. 
This filtering step is essential for managing complexity and targeting specific object groups within the scene.
\subsection{Object Querying by Relation (OQR)}
This group includes two core functions: \texttt{relate()} and \texttt{relate\_agent()}.
These functions are designed to retrieve objects based on spatial relationships with either a reference object or the agent itself.
This capability is essential for interpreting natural language expressions involving relative positioning, such as ``the cup on the table'' or ``the chair to my left''.
\begin{tcolorbox}[colback=gray!10,colframe=gray!80,title=API Documentation, breakable]
\begin{Verbatim}[breaklines=true, breaksymbol={}]
relate(object_set: Set[ObjectAttribute], reference_object: ObjectAttribute, relation: str) -> Set:
    """
    Returns a set of objects that are related to the reference_object by the relation.

    Examples
    --------
    >>> # Find objects on top of the table on my left
    >>> objects_on_table = set()
    >>> for table in table_left_set:
    >>>     objects_on_table.update(relate(object_set=object_set, reference_object=table, relation="on"))

    >>> # Determine what objects are on top of the table
    >>> objects_on_table_category = []
    >>> for obj in objects_on_table:
    >>>     objects_on_table_category.append(obj.category)
    >>> print(f"Objects on top of the table on my left: {objects_on_table_category}")
    Objects on top of the table on my left: ['book', 'tray']
    """
\end{Verbatim}
\end{tcolorbox}
Specifically, the \texttt{relate()} function takes a reference object and a relation descriptor (such as ``on,'' ``left,'' or ``behind'') and returns all objects in the input set that satisfy this relation with respect to the reference object. 
\begin{tcolorbox}[colback=gray!10,colframe=gray!80,title=API Documentation, breakable]
\begin{Verbatim}[breaklines=true, breaksymbol={}]
relate_agent(object_set: Set[ObjectAttribute], relation: str) -> Set:
    """
    Returns a set of objects that are related to the agent(you) by the relation.

    Examples
    --------
    >>> # Find the table on my left
    >>> table_left_set = relate_agent(object_set=table_set, relation="left")
    """    
\end{Verbatim}
\end{tcolorbox}
Similarly, \texttt{relate\_agent()} focuses on relations between objects and the agent itself.
Uses the agent itself as the reference point and retrieves objects that lie in the specified spatial relation relative to the agent. 

Collectively, these functions provide essential spatial reasoning capabilities, enabling the system to resolve object references in natural language queries within complex 3D environments by leveraging relative spatial relationships.
\subsection{Object Information Querying (OIQ)}
The Object Information Querying (OIQ) category comprises functions that retrieve detailed information about individual objects, including their spatial relationship to other objects or to the agent, as well as their physical and semantic properties.
These functions enable fine-grained understanding and reasoning about the scene by providing access to attributes such as position, shape, color, material, and state.
\begin{tcolorbox}[colback=gray!10,colframe=gray!80,title=API Documentation, breakable]
\begin{Verbatim}[breaklines=true, breaksymbol={}]
query_relation(object: ObjectAttribute, reference_object: ObjectAttribute, candidate_relations: Optional[List[str]]=["left", "right", "front", "back"]) -> List:
    """
    Returns a list of allcentric relations between the object and the reference_object.
    If `candidate_relations` is provided, only relations in the `candidate_relations` list will be returned.

    Examples
    --------
    >>> relation = query_relation(object=chair, reference_object=table)
    >>> print(f"The chair is in the direction of {' '.join(relation)} to the table")
    The chair is in the direction of left front to the table

    >>> relation = query_relation(object=chair, reference_object=table, candidate_relations=["left", "right"])
    >>> print(f"The chair is on the {' '.join(relation)} of the table")
    The chair is on the left of the table
    """
\end{Verbatim}
\end{tcolorbox} 
The \texttt{query\_relation()} function is used to determine the allocentric spatial relationship between two objects in a scene—that is, it describes how a target \texttt{object} is positioned relative to a \texttt{reference\_object}, independent of the agent’s viewpoint.
It returns a list of directional relations such as ``left,'' ``right,'' ``front,'' or ``back'' that describe this spatial configuration. 
By default, the function evaluates all four of these directions, but user can restrict the output to a custom subset by specifying a \texttt{candidate\_relations} list.
\begin{tcolorbox}[colback=gray!10,colframe=gray!80,title=API Documentation, breakable]
\begin{Verbatim}[breaklines=true, breaksymbol={}]
query_relation_agent(object: ObjectAttribute, candidate_relations: Optional[List[str]]=["left", "right", "front", "back", "o'clock"]) -> List:
    """
    Returns a list of allcentric relations between the object and the agent(you).
    If `candidate_relations` is provided, only relations in the `candidate_relations` list will be returned.

    Examples
    --------
    >>> # Decide which direction I should go to reach the table
    >>> direction = query_relation_agent(object=table)
    >>> print(f"Direction of the table relative to my current position: {direction}")
    >>> print(f"I should go {' '.join(direction)} to reach the table.")
    Direction of the table relative to my current position: ['left', 'back']
    I should go left back to reach the table.

    >>> # Decide whether the table is in front of me or behind
    >>> direction = query_relation_agent(object=table, candidate_relations=["front", "behind"])
    >>> print(f"Direction of the table relative to my current position: {' '.join(direction)}")
    Direction of the table relative to my current position: behind
    """
\end{Verbatim}
\end{tcolorbox}
The \texttt{query\_relation\_agent()} function determines the egocentric spatial relation between a specified object and the agent (i.e., ``you''), based on the agent’s current position and orientation in the environment.
It returns a list of direction labels such as ``left,'' ``right,'' ``front,'' ``back,'' or ``o'clock'' to describe where the object lies relative to the agent.
These labels help the agent interpret and navigate spatial environments from its own perspective.
Similarly, users can narrow down the output to a specific subset by providing a custom \texttt{candidate\_relations} list. 
\begin{tcolorbox}[colback=gray!10,colframe=gray!80,title=API Documentation, breakable]
\begin{Verbatim}[breaklines=true, breaksymbol={}]
query_attribute(object: ObjectAttribute, attribute_type: str, candidate_attribute_values: Optional[List[str]]) -> Union[List[float], float, str]:
    """
    Returns the attribute of the object.
    `attribute_type` must be chosen from the following list: ["lwh", "distance", "color", "shape", "material"].
    If `candidate_attribute_values` is provided, only values in the `candidate_attribute_values` list will be returned.

    Examples
    --------
    >>> lwh = query_attribute(object=object, attribute_type="lwh") # unit: meter. length, width, height of the object bounding box (unit: meter). Can be used to compute the length(lwh[0]), width(lwh[1]), height(lwh[2]), area(lwh[0]*lwh[1]) and volume(lwh[0]*lwh[1]*lwh[2]) of the object. Helpful for deciding the size of the object.
    >>> print(lwh)
    [0.68883693 0.29695976 0.17185348]

    >>> distance = query_attribute(object=object, attribute_type="distance") # unit: meter. Helpful for getting the distance of an object from the agent(you). Can be used to compare which object is closer or farther to the agent(you).
    >>> print(distance)
    2.3456789

    >>> # Determine whether the color of the object is brown, black or red
    >>> color = query_attribute(object=object, attribute_type="color", candidate_attribute_values=["brown", "black", "red"])
    >>> print(color)
    brown

    >>> # Determine whether the shape of the object is round, square or rectangular
    >>> shape = query_attribute(object=object, attribute_type="shape", candidate_attribute_values=["round", "square", "rectangular"])
    >>> print(shape)
    rectangular

    >>> # Determine whether the material of the object is wood or metal
    >>> material = query_attribute(object=object, attribute_type="material", candidate_attribute_values=["wood", "metal"])
    >>> print(material)
    wood
    """
\end{Verbatim}
\end{tcolorbox}
The \texttt{query\_attribute()} function extracts particular properties of an object such as its color, shape, material, size dimensions, or distance from the agent. When candidate attribute values are provided, it selects the most fitting attribute value for the object. 
\begin{tcolorbox}[colback=gray!10,colframe=gray!80,title=API Documentation, breakable]
\begin{Verbatim}[breaklines=true, breaksymbol={}]
query_state(object: ObjectAttribute, candidate_states: List[str]) -> str:
    """
    Returns the state of the object.

    Examples
    --------
    >>> state = query_state(object=object, candidate_states=["neat", "messy"])
    >>> print(state)
    neat
    """
\end{Verbatim}
\end{tcolorbox}
The \texttt{query\_state()} function determines the current state of an object (for example, ``neat'' or ``messy'') from a list of candidate states.

Together, these functions facilitate a rich and nuanced understanding of the environment, supporting sophisticated reasoning and interaction based on both physical and semantic object properties.

\section{Prompt Construction}
\subsection{Prompts for Task Execution}

The prompts below are provided at the very beginning, along with the API documentation to guide the LLM in the process of solving situated reasoning tasks through iterative steps involving reasoning-action-feedback observation. 
These prompts establish a clear protocol for the LLM to reason about the scene, generate plans and programs, and ultimately produce concise final answers. 
The structure ensures that the LLM systematically explores the environment using available APIs and applies commonsense reasoning to infer missing information.

Specifically, the task execution prompt instructs the LLM to alternate between reasoning steps (Thought), decision making on the next step (Action), and supplying the corresponding input (Action Input). Two main action types are supported:
\begin{tcolorbox}[colback=gray!10,colframe=gray!80,title=Prompt, breakable]
\begin{Verbatim}[breaklines=true, breaksymbol={}]
You are a smart embodied agent. Use your coding and common sense reasoning skills to solve a question answering task with interleaving Thought, Action, Observation steps. Given your situation and question, use the following format to solve the task:
Thought: Answer the question by reasoning about scene and your situation. If you need further information about the objects in the scene (e.g. spatial relationship), generate a plan step by step and implement it in a program.
Action: The action to take, should be one of [Final Answer, Program].
Action Input:
(1) For Final Answer, return the answer to the question with NO MORE THAN 3 words.
(2) For Program, generate a Python program according to your thought to help you understand the scene.
\end{Verbatim}
\end{tcolorbox}
For the final answer, upon receiving the output of the executable code, the LLM verifies its conclusion and provides a concise answer—no more than three words—in the \texttt{Action Input} field, based on the observed results and following the format below.
\begin{tcolorbox}[colback=gray!10,colframe=gray!80,title=Prompt, breakable]
\begin{Verbatim}[breaklines=true, breaksymbol={}]
Valid format for Final Answer
-----------------------------
Thought: Your reasoning process for the final answer. 
Action: Final Answer
Action Input: Your final answer with NO MORE THAN 3 words. (Use your common sense reasoning skills to infer missing information and give a specific final answer.)
\end{Verbatim}
\end{tcolorbox}
For the initial reasoning step or in cases where an error occurs during code execution, the LLM is required to perform a \texttt{Program} action by generating a well-structured Python script. 
This program should be used to query and analyze objects within the scene. 
The code must include print statements to display the values of relevant variables, enabling observation and interpretation of the results.
\begin{tcolorbox}[colback=gray!10,colframe=gray!80,title=Prompt, breakable]
\begin{Verbatim}[breaklines=true, breaksymbol={}]
Valid format for Program
------------------------
Thought: Your plan for further information about the objects in the scene.
Action: Program
Action Input:

Python
YOUR PROGRAM (Use
print(variable_value_you_want_to_know)
to display the value of a variable.)
\end{Verbatim}
\end{tcolorbox}
In addition, an extra \texttt{Tips} section is provided to help ensure that the LLM adheres as closely as possible to the specified generation requirements and to prevent common generation errors.
\begin{tcolorbox}[colback=gray!10,colframe=gray!80,title=Prompt, breakable]
\begin{Verbatim}[breaklines=true, breaksymbol={}]
**Tips**
1. ALWAYS adhere to the valid output format.
2. Pass the correct parameter types to the function.
3. Try to infer missing information using your commonsense reasoning skills.
4. Use 
print(variable_value_you_want_to_know)
 to display the value of a variable. Otherwise, you would get nothing from the observation.
5. Consider all the objects in a set, instead of querying only one object's attribute.
6. Return the Final Answer with NO MORE THAN 3 words.
\end{Verbatim}
\end{tcolorbox}
\subsection{Prompt for Dataset Extension}
During the dataset expansion stage, the original questions from the SQA3D dataset are enriched and made more challenging with the assistance of an LLM.
The specific prompt used for data generation is shown below.
\begin{tcolorbox}[colback=gray!10,colframe=gray!80,title=Prompt, breakable]
\begin{Verbatim}[breaklines=true, breaksymbol={}]
You are a robot specialized in 3D scenes. Your position is as follows:
```
{situation}
```
Here is a series of existing questions and answers in the this scene.
```
{previous_questions}
```
You need to combine these questions and answers to construct {num_questions} more complex questions to increase the reasoning difficulty.
IMPORTANT: 
1. you need to reply in the following json format, just a list of questions. no other words.
2. {num_questions} is the number of questions you need to generate.
<return_format>
```json
[
    ...
    {{
        "answer": "three",
        "question": "How many chairs are immediately to the left?"
    }}
    ...
]
```
\end{Verbatim}
\end{tcolorbox}
In this prompt, the LLM is instructed to assume the role of a robot specialized in understanding 3D environments. 
The input includes the agent’s current situation (\texttt{situation}), as well as a list of previously asked questions and their corresponding answers (\texttt{previous\_questions}) within the same scene. 
The model’s task is to synthesize this information and generate a specified number (\texttt{num\_questions}, that is half of SQA3D's number in our experiment) of new questions that require deeper reasoning, such as combining spatial relations, counting, and comparative analysis across multiple objects or perspectives.

To ensure consistency and machine-readability, the model is required to return the output strictly in a predefined JSON format, where each entry includes a question and its corresponding answer. 
No explanatory text or additional commentary is allowed. 
This structured format supports automatic integration of the generated data into existing datasets and facilitates downstream model training.

The prompt serves to enhance the original SQA3D dataset by introducing higher-order reasoning tasks, encouraging the model to infer, compare, and synthesize information beyond surface-level object attributes. 
\section{Additional Results}
\subsection{Further Analysis of the Correct Results}
% \begin{figure*}[t]
%     % \vspace{-0.3cm}
%     \centering
%     \includegraphics[width=\columnwidth]{Figures/success.png}
%     \vspace{-0.7cm}
%     \caption{Examples of Success Cases}
%     \label{fig:com}
%     % \vspace{-0.3cm}
% \end{figure*}
\subsubsection{Qualitative Results}
In the visualization experiments, we intentionally select several representative examples to demonstrate the model’s ability to invoke APIs accurately, its deep comprehension of 3D environments, and its adaptability in addressing a wide range of spatial and situational reasoning challenges.

\begin{figure*}[t]
    % \vspace{-0.3cm}
    \centering
    \includegraphics[width=\columnwidth]{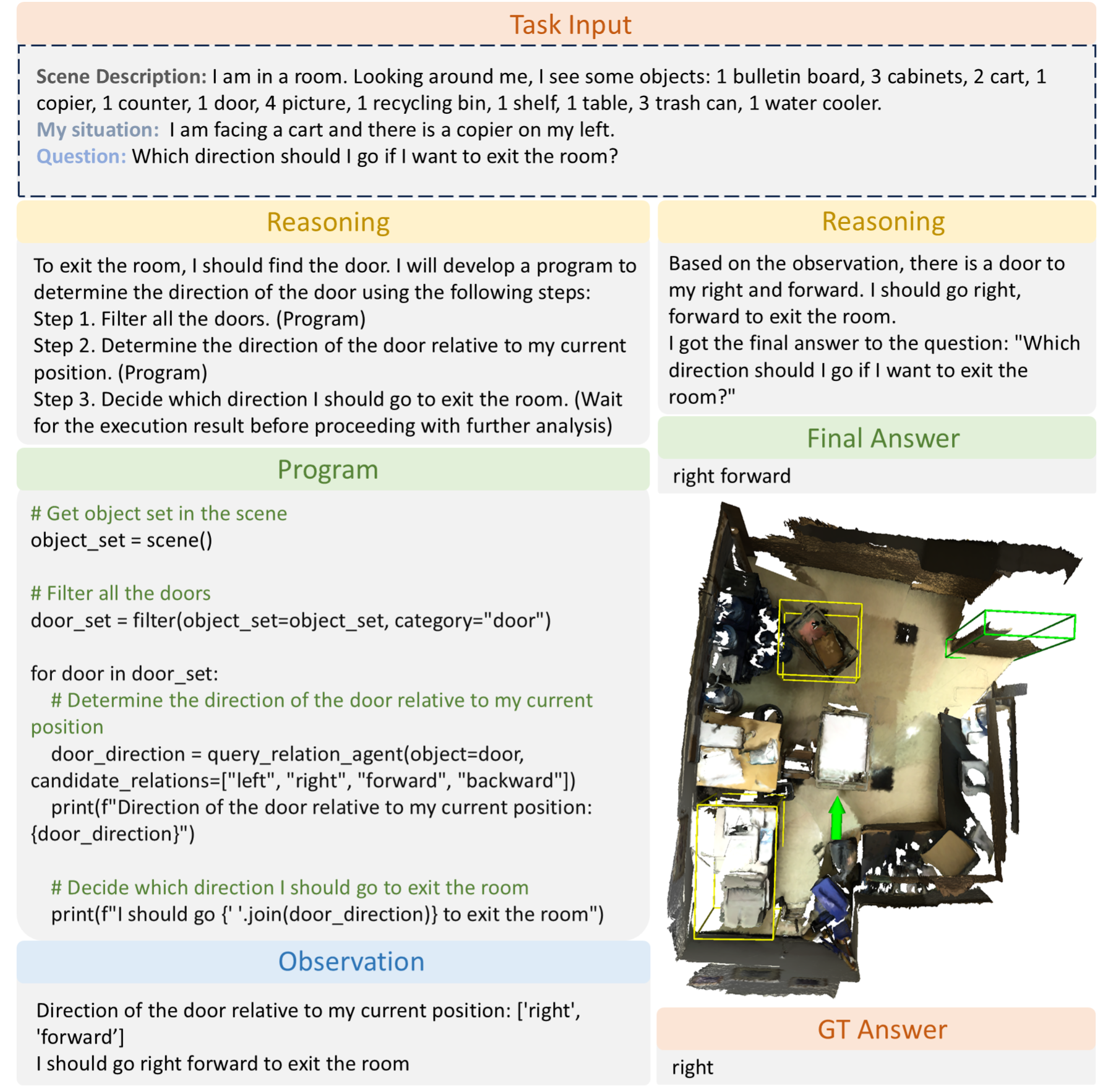}
    % \vspace{-0.7cm}
    \caption{\textbf{Success on Spatial Localization and Navigation Reasoning.} The generated code is free of any redundancy, showcasing the model’s precise understanding and efficient utilization of the provided APIs.}
    \label{fig:1}
    % \vspace{-0.3cm}
\end{figure*}

As illustrated in Fig.~\ref{fig:1}, the model demonstrates strong capabilities in spatial localization and navigation reasoning. 
The success in this task can be largely attributed to the model’s robust understanding and proficient use of the provided APIs. 
In this scenario, the model first identifies the essential condition for exiting the room—namely, the presence of a door. 
It then accurately invokes the \texttt{filter()} function to identify the door object within the room, followed by a precise use of the \texttt{query\_relation\_agent()} function to determine the spatial relationship between the agent and the door, ultimately inferring the correct direction for exiting. 
The API calls generated by the model are concise and purposeful, free of redundancy, clearly demonstrating its precise comprehension and effective use of the available API set.

\begin{figure*}[t]
    % \vspace{-0.3cm}
    \centering
    \includegraphics[width=\columnwidth]{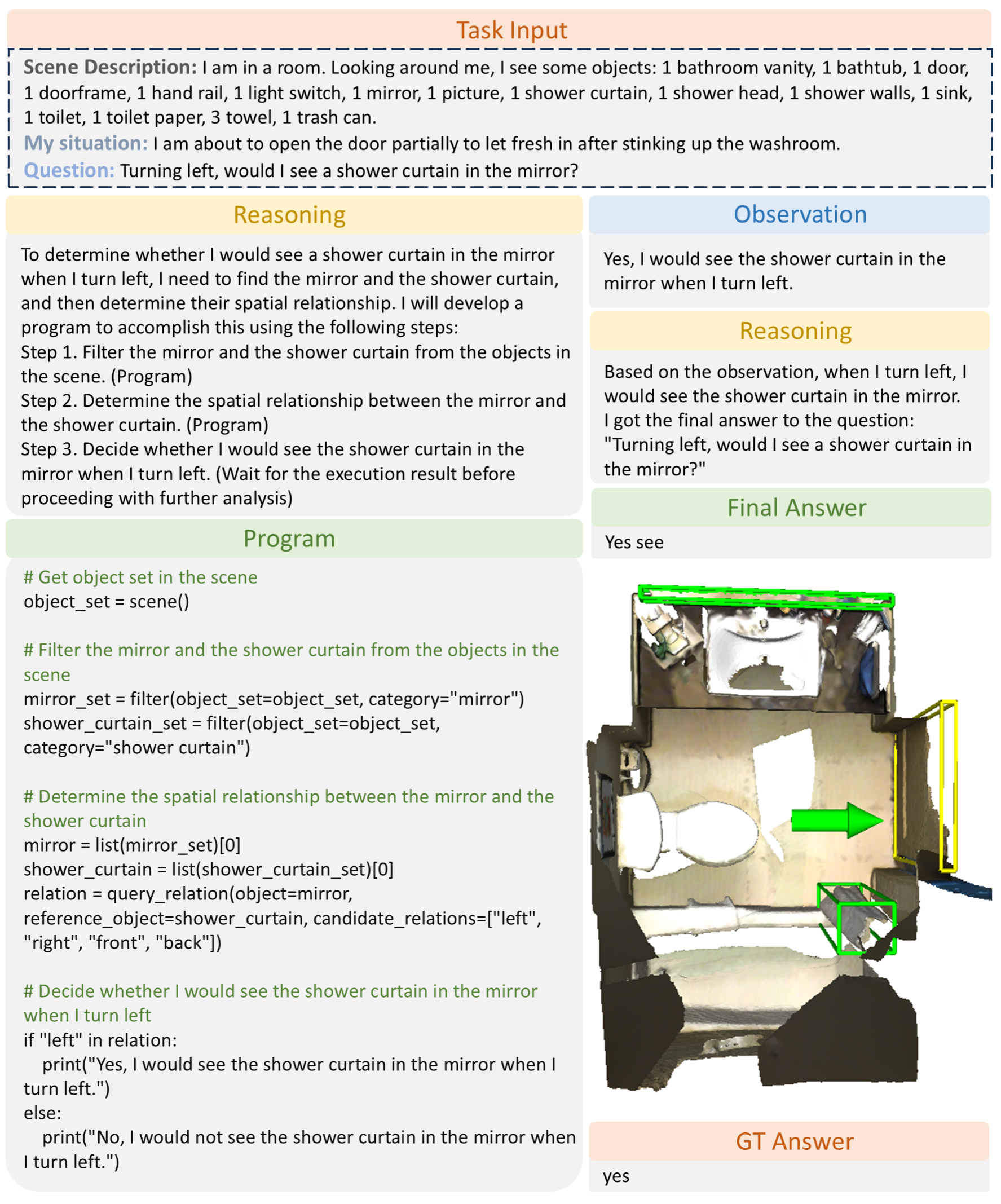}
    % \vspace{-0.7cm}
    \caption{\textbf{Success on Object Understanding and Detail Reasoning.} Our model combines the results obtained through API calls with its internal knowledge to develop a deeper understanding of the scene.}
    \label{fig:2}
    % \vspace{-0.3cm}
\end{figure*}

In addition to precise API invocation, the model also demonstrates a deep understanding of the scene layout and the functional roles of objects within it.
As illustrated in Fig.~\ref{fig:2}, the model accurately understands the reflective property of the mirror in the scene and determines whether the agent can see the curtains in the mirror based on the spatial relationships among the mirror, the curtains, and the agent. 
The model not only grasps the functional role of the mirror as an object but also leverages deeper knowledge about visual reflection and perspective to reason about what is visible from the agent’s viewpoint.
This indicates that the model's understanding of the scene goes beyond surface-level descriptions gained from API results—it integrates its internal knowledge with spatial context to achieve a more profound comprehension of the environment.

\begin{figure*}[t]
    % \vspace{-0.3cm}
    \centering
    \includegraphics[width=0.75\columnwidth]{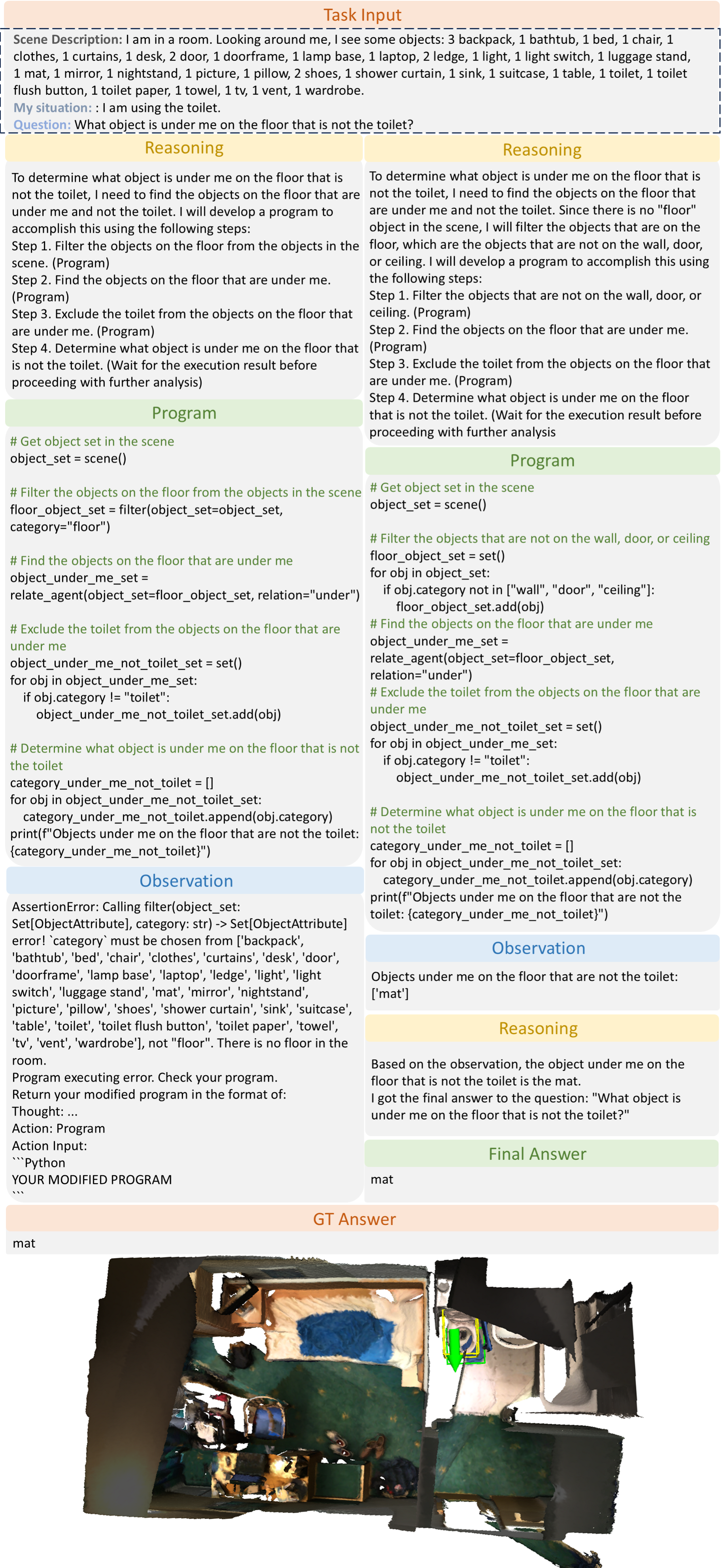}
    % \vspace{-0.7cm}
    \caption{\textbf{Example of Self-Correction Capability.} Our model is capable of self-correcting errors by reassessing its reasoning process and generating revised code accordingly.}
    \label{fig:3}
    % \vspace{-0.3cm}
\end{figure*}

Our model also exhibits strong self-correction capabilities, extending beyond merely fixing code errors to include self-correction in logical reasoning. 
As illustrated in Fig.~\ref{fig:3}, during its initial reasoning, the model directly searched for objects labeled as ``floor'' using straightforward logic. 
However, upon encountering an error indicating the absence of a ``floor'' label, the model revised its approach in the second attempt by modifying the filtering logic. 
Instead of relying on the nonexistent label, it applied directional reasoning to exclude objects located on wall, door, and ceiling, thereby isolating those on the floor. 
This demonstrates the model’s flexible logical reasoning ability to dynamically correct its own reasoning chain and achieve the final goal through alternative strategies.

\subsubsection{Distribution of Communication Rounds}

In this experiment, we compare the distribution of communication rounds required to obtain the correct answers between our method and LLM-TPC~\cite{qingrong2024llm-tpc}.

\begin{figure*}[t]
    % \vspace{-0.3cm}
    \centering
    \includegraphics[width=\columnwidth]{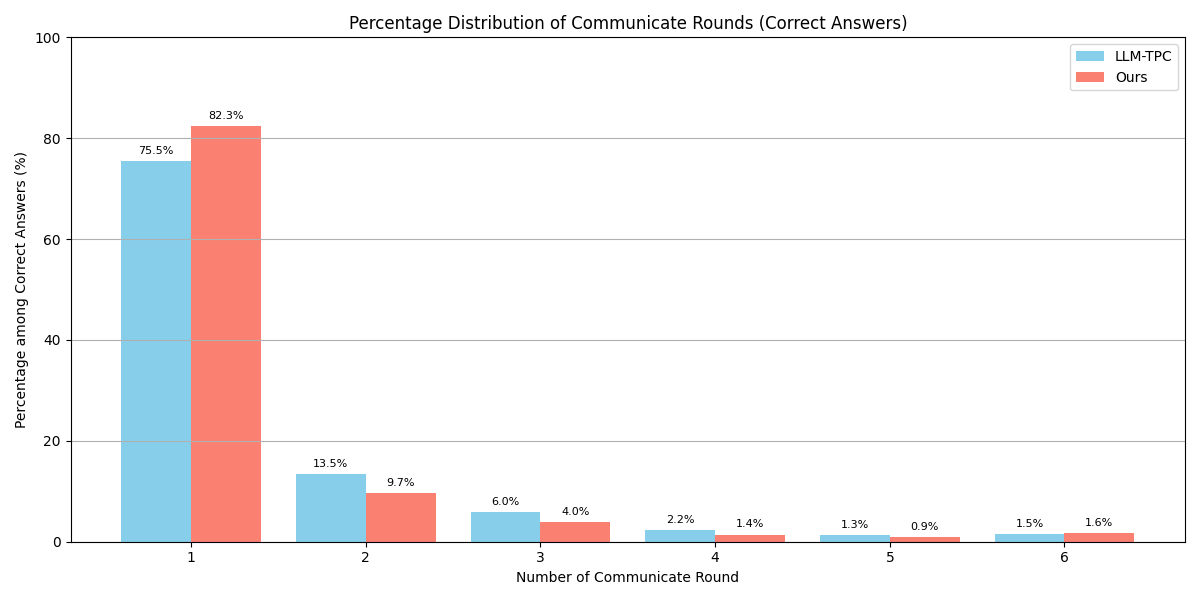}
    \vspace{-0.7cm}
    \caption{\textbf{Distribution of Communication Rounds in Correct Answers.} Our model demonstrates a significant improvement in generating correct answers within a single generation round.}
    \label{fig:4}
    % \vspace{-0.3cm}
\end{figure*}

As shown in Fig.~\ref{fig:4}, 82.3\% of our model's results achieve the correct code and answer in the first generation round, which is nearly 7\% higher than that of LLM-TPC.
This indicates that the integration of SFT and DPO into our model has significantly enhanced the overall framework’s understanding of tasks and reasoning processes, while also improving the executability of the generated code. 
As a result, the model demonstrates a substantial improvement in one-shot generation accuracy.

In later rounds of correction, our DeepThink3D consistently shows a lower proportion of cases requiring 2–5 communication rounds compared to LLM-TPC, largely because more of our results are already correct in the first round. 
% While it's worth noting that, even in the 6-round communication setting, our method still matches or slightly exceeds LLM-TPC in terms of success rate.
Although our model has demonstrated a clear advantage in obtaining correct answers in just one round, an interesting phenomenon is worth noting: there remains a small proportion of cases—slightly higher than LLM-TPC by 0.1\%—where our DeepThink3D requires up to six rounds to arrive at the correct answer.
Alternatively, this may suggest that our model is capable of persistently refining its reasoning process over multiple rounds, even when the initial attempts are unsuccessful, showcasing a strong capacity for iterative problem-solving and resilience in complex scenarios.

\subsection{Analysis of the Incorrect Results}
In this experiment, 30 tasks with incorrect answers were randomly selected for error type analysis. 
The statistical results indicate that the errors in our results can be mainly categorized into the following five types:
\begin{itemize}
\item \textbf{API Error}: This type of error is primarily caused by discrepancies between the output of API function calls and the actual scene, resulting in inconsistencies with the final answer or affecting critical intermediate reasoning steps;
\item \textbf{Annotation Error}: This issue mainly arises from the rigid matching of object labels; for example, ``file cabinet'' and ``cabinet'' are treated as two distinct categories in the API’s category parameter;
\item \textbf{Logic Error}: This type of problem is caused by logical reasoning errors of the LLM, often due to oversimplified thinking or discrepancies between the LLM’s reasoning patterns and human intuition;
\item \textbf{Judgement Error}: This type of error appears in the final answer’s output format; although the semantic meaning of the generated answer is correct, it is mistakenly marked as incorrect due to the soft matching failing to align it with the ground truth answer;
\item \textbf{Task Understanding Error}: This type of error is caused by the model’s misunderstanding of the question’s intent or by inherent ambiguity within the question itself.

\end{itemize}

\begin{figure*}[t]
    % \vspace{-0.3cm}
    \centering
    \includegraphics[width=\columnwidth]{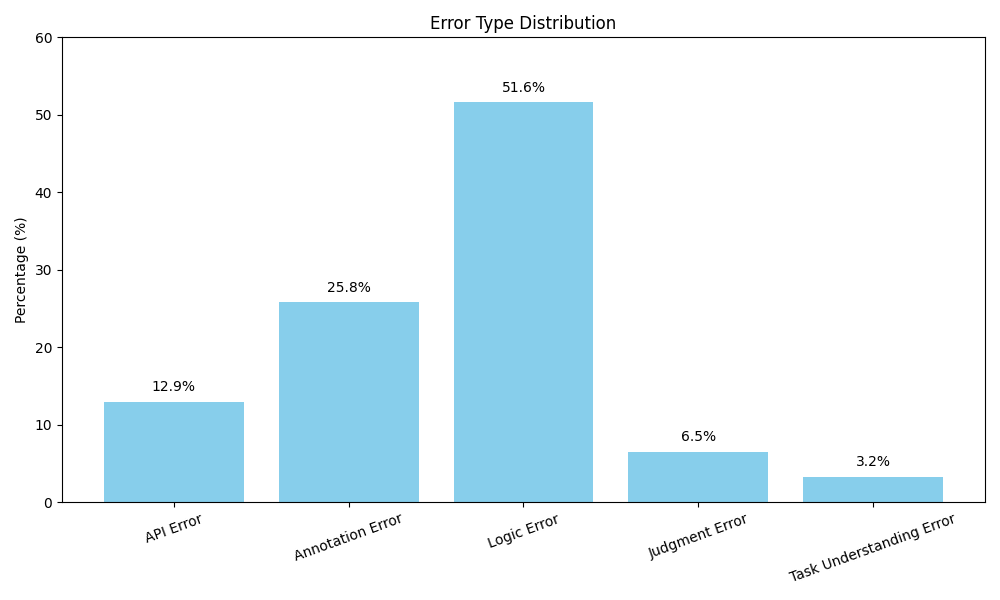}
    \vspace{-0.7cm}
    \caption{Error Type Distribution in Incorrect Answers.}
    \label{fig:5}
    % \vspace{-0.3cm}
\end{figure*}

As shown in the Fig.~\ref{fig:5}, the errors in our results are mainly composed of Logic Errors, Annotation Errors, and API Errors. 
Judgement Errors and Task Understanding Errors are relatively rare, accounting for only about 1 or 2 out of the 30 sampled cases.
Here, several representative examples have been selected to illustrate these common types of errors.

\begin{figure*}[t]
    % \vspace{-0.3cm}
    \centering
    \includegraphics[width=\columnwidth]{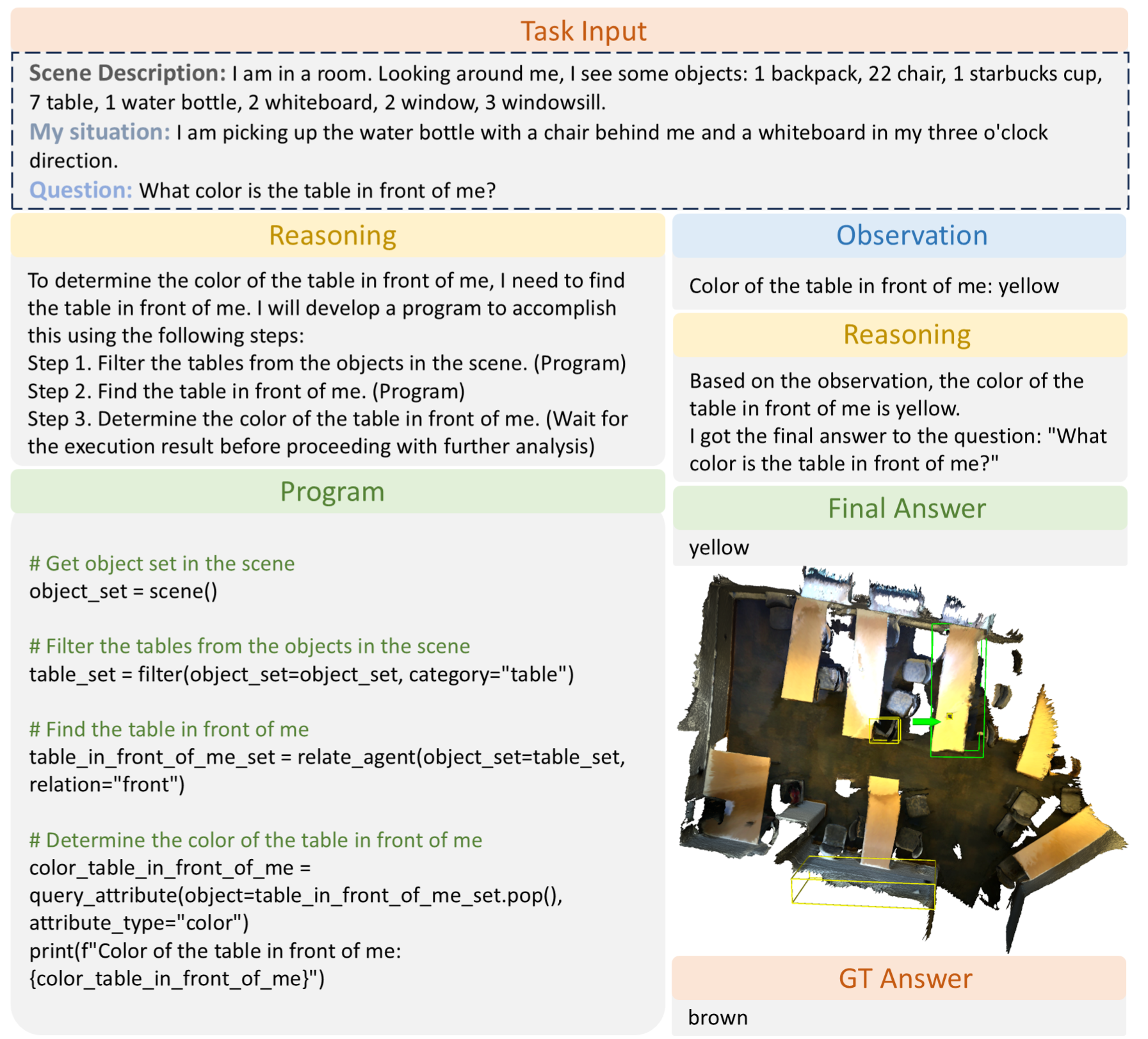}
    % \vspace{-0.7cm}
    \caption{Example of \textbf{API Error}.}
    \label{fig:6}
    % \vspace{-0.3cm}
\end{figure*}

The example shown in Fig.~\ref{fig:6} illustrates an \textbf{API error} caused by the \texttt{query\_attribute()} function. In this case, the color of the desk is mistakenly identified as yellow instead of brown. This error can be traced back to a discrepancy between the OpenShape~\cite{liu2023openshape} module’s extracted object attribute features and the actual values, which ultimately led to the incorrect result.
Within the current model framework, addressing these API output errors caused by modules can involve incorporating post-processing and validation mechanisms. However, since these errors typically occur at the fundamental level of object attributes, fully eliminating them without manual verification remains challenging. Nonetheless, we can look forward to the development of more accurate modules in the future to replace the existing ones.

\begin{figure*}[t]
    % \vspace{-0.3cm}
    \centering
    \includegraphics[width=\columnwidth]{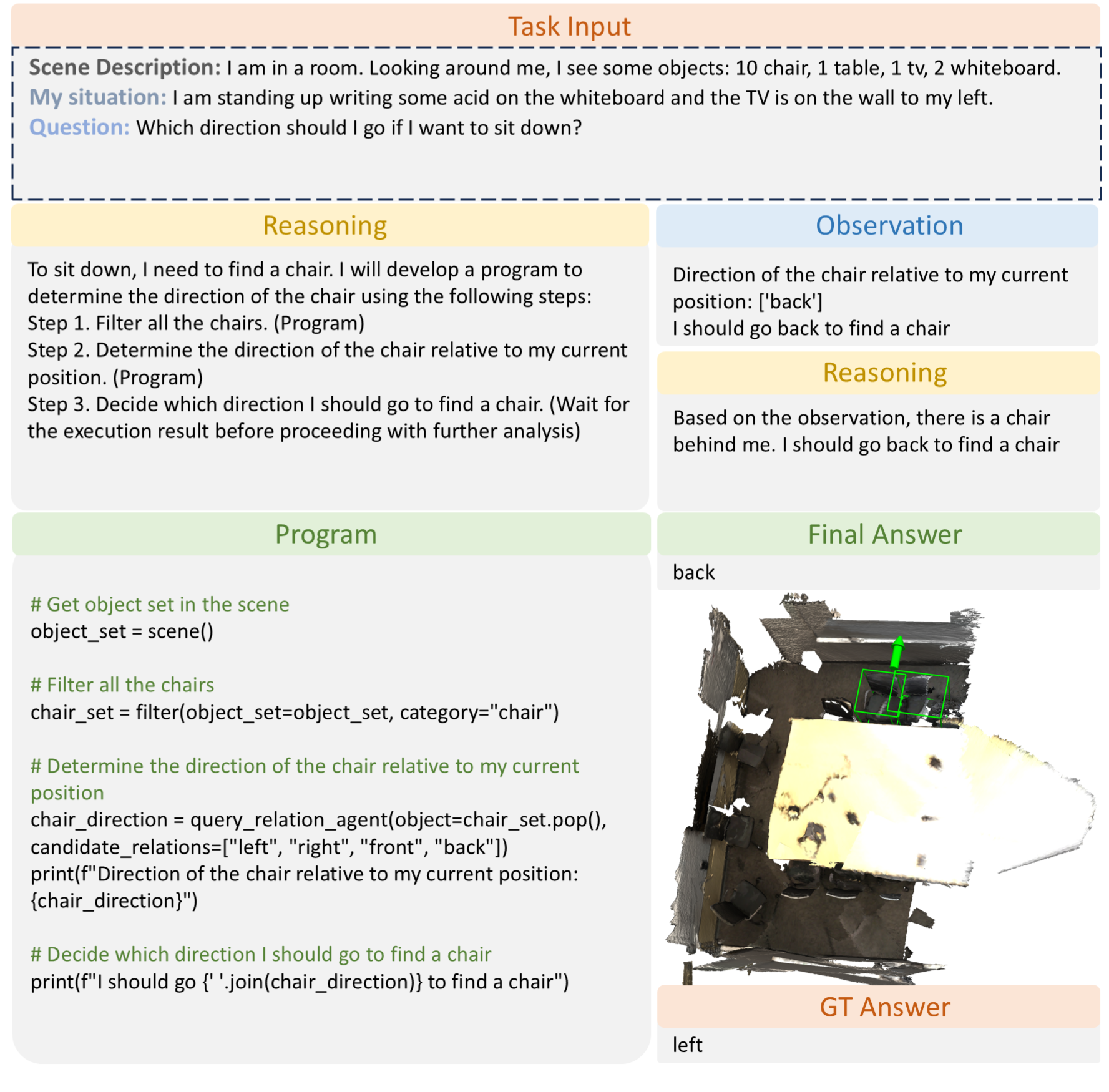}
    % \vspace{-0.7cm}
    \caption{Example of \textbf{Annotation Error}.}
    \label{fig:7}
    % \vspace{-0.3cm}
\end{figure*}

For \textbf{annotation errors}, these types of errors mainly include GT object segmentation not being fully marked, point cloud objects being incomplete and not having enough details, and GT answers leading to a mistake. As shown in Fig.~\ref{fig:7}, the agent requests to find a chair to sit down. From the current viewpoint, two chairs are behind it, allowing the agent to sit down directly by moving backward, while the GT answer shows a left direction. To address these errors, it remains necessary for us to strive to construct a dataset with higher precision and improve the annotation accuracy.

\begin{figure*}[t]
    % \vspace{-0.3cm}
    \centering
    \includegraphics[width=\columnwidth]{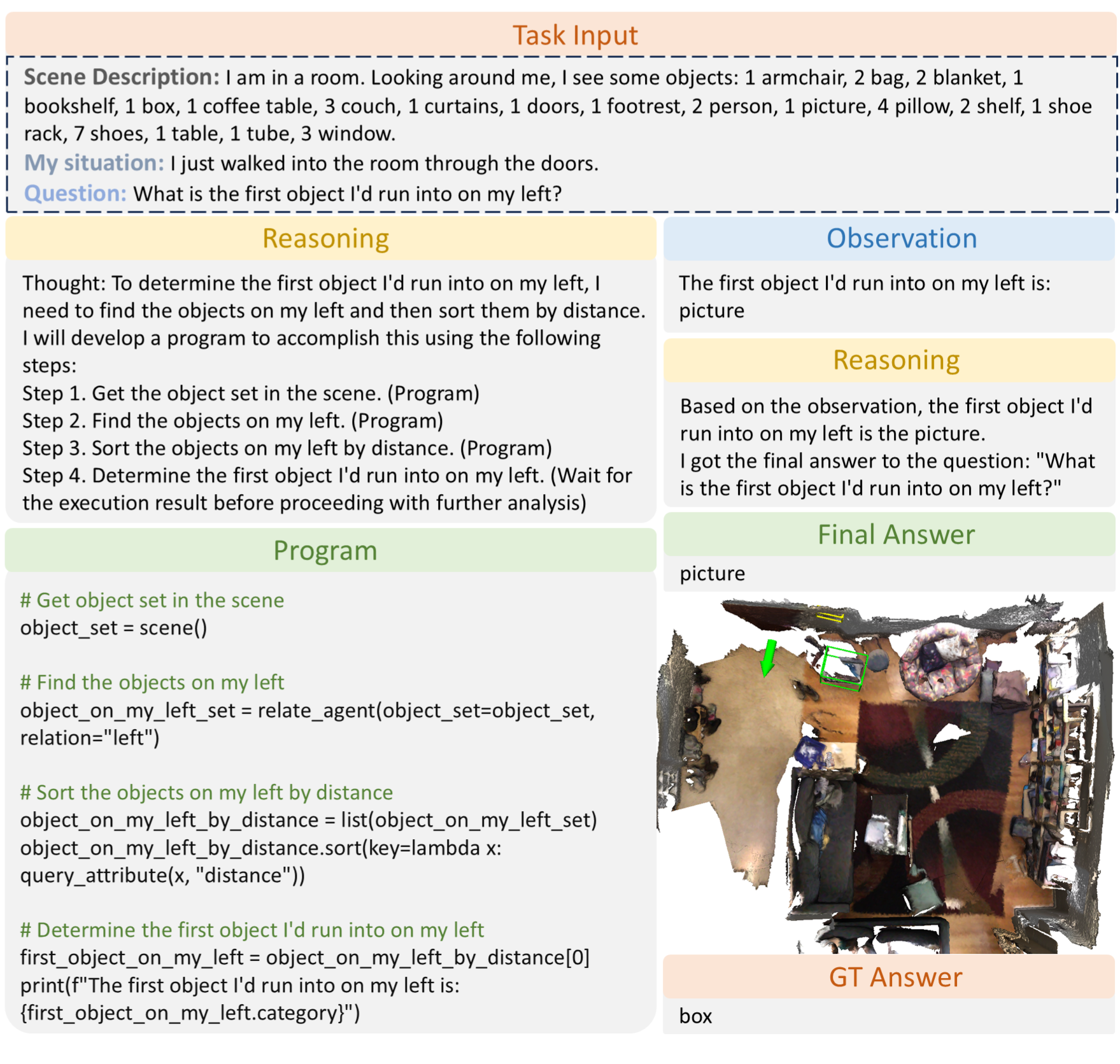}
    % \vspace{-0.7cm}
    \caption{Example of \textbf{Logic Error}.}
    \label{fig:8}
    % \vspace{-0.3cm}
\end{figure*}

Finally, the \textbf{logic errors} represent one of the major challenges we need to address. 
Some of these errors occur because the LLM’s reasoning chain is overly simplistic. 
This issue can be mitigated to some extent through further model training and fine-tuning — in fact, our incorporation of SFT and DPO for joint training of the LLM has already helped alleviate some of these problems, though there is still room for improvement. 
However, other errors arise due to inherent differences between the LLM’s reasoning patterns and human cognitive habits. 
For example, in the case shown in Fig.~\ref{fig:8}, the LLM fails to fully grasp the semantic nuance of “run into” in human context. 
To the model, “run into” simply means the closest or first object encountered. 
But from a human perspective, “run into” usually excludes objects like the “picture” given in the answer — which is not actually along the path of movement and does not constitute an obstacle. 
This mismatch in interpretation leads to the deviation in the final answer.
To address such issues, approaches like ours—which involve task-specific fine-tuning and optimition of the LLM for 3D Situated Reasonin—can achieve some improvements. However, a significant portion of the solution still depends on further advancements in aligning the LLM’s understanding more closely with true human semantic comprehension, which requires our continued efforts and further research.

% \newpage
% \clearpage
% \bibliographystyle{unsrt}
% \bibliography{ref}

% \end{document}

\end{document}